%% file: main.tex
\documentclass[10pt,twocolumn,letterpaper]{article}

\usepackage{blindtext}
\usepackage{multirow, makecell}
\usepackage{bm}
\usepackage{bbm} 
\usepackage{comment}
\usepackage{algorithm}
\usepackage{algorithmic}
\usepackage{array} 
\usepackage[dvipsnames]{xcolor}
\usepackage[safe]{tipa} 
\usepackage{color, colortbl}
\usepackage[accsupp]{axessibility}
\usepackage[misc,geometry]{ifsym} 
\usepackage{tabularx} 
\usepackage{transparent}
\makeatletter
\renewcommand\paragraph{
  \@startsection{paragraph} 
  {4} 
  {\z@} 
  {.5em \@plus1ex \@minus.2ex} 
  {-1.5em} 
  {\normalfont\normalsize\bfseries} 
}
\makeatother

\makeatletter
\def\@fnsymbol#1{\ensuremath{\ifcase#1\or \textsuperscript{~\Letter}\or \ddagger\or
   \mathsection\or \mathparagraph\or \|\or **\or \dagger\dagger
   \or \ddagger\ddagger \else\@ctrerr\fi}}
\makeatother

\definecolor{tabhighlight}{HTML}{e5e5e5}
\definecolor{citecolor}{HTML}{0071bc}
\definecolor{Gray}{gray}{0.95}

\newcommand{\gray}[1]{\textcolor{gray}{{#1}}}

\newcommand{\white}[1]{\color[HTML]{FFFFFF}{{#1}}}

\newcolumntype{g}{>{\columncolor{Gray}}c}

\usepackage{cvpr}              

\input{preamble}

\definecolor{cvprblue}{rgb}{0.21,0.49,0.74}
\usepackage[pagebackref,breaklinks,colorlinks,citecolor=cvprblue]{hyperref}

\def\paperID{877} 
\def\confName{CVPR}
\def\confYear{2024}

\title{OVMR: Open-Vocabulary Recognition with Multi-Modal References}

\author{
Zehong~Ma\textsuperscript{\rm 1} \quad
Shiliang~Zhang\textsuperscript{\rm 1}\footnotemark[2] \quad
Longhui~Wei\textsuperscript{\rm 2} \quad
Qi~Tian\textsuperscript{\rm 2} \\
\textsuperscript{\rm 1}National Key Laboratory for Multimedia Information Processing, \\ School of Computer Science, Peking University, \textsuperscript{\rm 2}Huawei Inc.
\\
{\tt\small zehongma@stu.pku.edu.cn, slzhang.jdl@pku.edu.cn, weilh2568@gmail.com, tian.qi1@huawei.com}
}

\begin{document}
\maketitle
\footnotetext[2]{Corresponding author.}
\input{sec/0_abstract}    
\input{sec/1_intro}

\input{sec/2_related_work}
\input{sec/3_method}
\input{sec/4_experiments}

\input{sec/5_conclusion}

\noindent\textbf{Acknowledgement} This work is supported in part by Natural Science Foundation of China under Grant No. U20B2052, 61936011, in part by the Okawa Foundation Research Award.

{
    \small
    \bibliographystyle{ieeenat_fullname}
    \bibliography{main}
}

\input{sec/X_suppl}

\end{document}

%% file: preamble.tex
\usepackage[dvipsnames]{xcolor}


%% file: sec/0_abstract.tex
\begin{abstract}

The challenge of open-vocabulary recognition lies in the model has no clue of new categories it is applied to. Existing works have proposed different methods to embed category cues into the model, \eg, through few-shot fine-tuning, providing category names or textual descriptions to Vision-Language Models. Fine-tuning is time-consuming and degrades the generalization capability. Textual descriptions could be ambiguous and fail to depict visual details. This paper tackles open-vocabulary recognition from a different perspective by referring to multi-modal clues composed of textual descriptions and exemplar images. Our method, named OVMR, adopts two innovative components to pursue a more robust category cues embedding. A multi-modal classifier is first generated by dynamically complementing textual descriptions with image exemplars. A preference-based refinement module is hence applied to fuse uni-modal and multi-modal classifiers, with the aim to alleviate issues of low-quality exemplar images or textual descriptions. The proposed OVMR is a plug-and-play module, and works well with exemplar images randomly crawled from the Internet. Extensive experiments have demonstrated the promising performance of OVMR, \eg, it outperforms existing methods across various scenarios and setups. Codes are publicly available at \href{https://github.com/Zehong-Ma/OVMR}{https://github.com/Zehong-Ma/OVMR}.


\end{abstract}

%% file: sec/1_intro.tex
\section{Introduction}
\label{sec:intro}

Open-vocabulary recognition aims to recognize unseen objects beyond the training set. It is challenging because the model has no clue of new categories in the testing set. Besides efforts on pre-training models having strong generalization capability~\cite{CLIP, ni2023continual, sun2023eva}, recent works have developed more lightweight strategies by embedding novel category clues into pre-trained backbone models~\cite{coop, khattak2023maple}. Among those works, a popular strategy is fine-tuning a generalizable model on a small task-specific dataset~\cite{coop, cocoop, wu2023cora}. This few-shot fine-tuning strategy is effective in optimizing task-specific parameters, but is time-consuming, inflexible, and degrades the generalization capabilities. 

\begin{figure}
    \centering
    \includegraphics[width=0.9\linewidth]{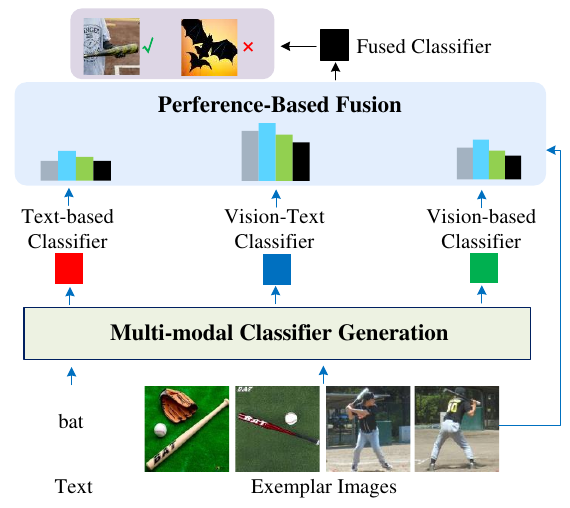}
    \caption{{Illustration of the pipeline of our OVMR. It refers to textual description and exemplar images to generate classifiers for novel categories. The textual description could be ambiguous and fail to depict visual details. The exemplar images show diversified qualities. OVMR effectively complements visual and textual features and fuses classifiers to alleviate issues of low-quality exemplar images or textual description.}
    }
    \label{fig1:intro}
    \vspace{-5mm}
\end{figure}

Another line of research is leveraging the strong generalization capability of Vision-Language Models (VLMs) by providing images or textual descriptions as clues of novel categories~\cite{zang2022open, kaul2023multi}. Some works take text embeddings extracted from textual descriptions as the classifier for novel categories~\cite{zhou2022detecting, wu2023cora}. Textual descriptions can be ambiguous and lack detailed descriptions to visual cues. For example, the word ``bat'' could refer to either a piece of sports equipment or an animal. Those issues scarify the discriminative power of resulting classifiers. Collecting exemplar images could be another option for providing category cues as shown in previous works~\cite{kaul2023multi, xu2023multi}. However, image samples could show diversified qualities easily affected by issues of domain gaps, cluttered backgrounds, \etc.

This paper tackles open-vocabulary recognition from a different perspective by referring to multi-modal clues composed
of textual descriptions and exemplar images. In other words, by feeding both text descriptions and exemplar images into the VLM, we aim to mine complementary cues of text and images to learn more robust classifiers for novel categories. As illustrated in Fig.~\ref{fig1:intro}, during this procedure, the text modality is expected to provide generalizable semantic cues, while the exemplar images are analyzed to extract visual details, which are critical to the discriminative power of resulting classifiers. To alleviate the negative effects of low-quality text or image exemplars, we also evaluate the performance of those uni-modal and multi-modal classifiers to adaptively generate the final classifier. We name the proposed method as OVMR.

As shown in Fig.~\ref{fig1:intro}, OVMR takes textual descriptions and multiple exemplar images depicting a novel category as input. It incorporates two modules to generate the final classifiers. The first module dynamically fuses visual exemplars and textual descriptions to generate the multi-modal classifier. Specifically, it utilizes a lightweight visual token generator to extract visual tokens from given exemplars. Subsequently, the language encoder adaptively fuses the visual and textual tokens by inferring the contextual relationships between them~\cite{liu2023llava,xuan2023pink}. This multi-modal classifier generation module is efficient thanks to its lightweight structure. It does not need to train class-specific parameters, hence ensuring its good generalizability and scalability to the classes in the wild. This module leads to two uni-modal classifiers, and one multi-modal classifier.

OVMR hence generates the final classifier by fusing the above three classifiers in Fig.~\ref{fig1:intro}. To alleviate the negative effects of low-quality classifiers, it presents a dynamic fusion strategy by evaluating their performance. As multiple exemplar images are provided, we leverage them as a validation set to test the performance of each classifier. The preference-based fusion module uses their performance as the clue of learning fusion weights. As shown in Fig.~\ref{fig1:intro}, this procedure simulates the testing stage by leveraging exemplar images as the testing set. It effectively guarantees the robustness of the final fused classifier.

We have conducted extensive experiments to test the performance of OVMR. As shown in experiments, it outperforms recent open-vocabulary methods by clear margins on 11 image classification datasets, and the LVIS object detection dataset. OVMR also performs better than closely related works that simply apply naive average fusion~\cite{kaul2023multi} and text-guided fusion~\cite{xu2023multi} for novel classifier generation. Our contributions can be summarized into three aspects:

\begin{itemize}
    \item We present a flexible plug-and-play module to embed clues of novel classes into VLMs to boost their capabilities in open-vocabulary recognition tasks. Complementing multi-modal clues brings substantial advantages over solely relying on vision or textual cues. 
    \item Our OVMR presents a novel pipeline to generate robust classifiers from two-modality inputs. It adaptively fuses text and vision cues to generate multi-modal classifiers, and further proposes a parameter-free fusion module to alleviate negative effects of the low-quality modality. 
    \item Extensive experiments demonstrate the superior performance of our method in both open-vocabulary classification and detection tasks, showcasing the potential of our method in open-vocabulary recognition.
\end{itemize}

%% file: sec/2_related_work.tex
\section{Related Work}
\label{sec:related}



This work is closely related to open-vocabulary classification and detection. This section briefly reviews recent works in those two lines and discusses our differences with them.

\subsection{Open-Vocabulary Classification}

Existing open-vocabulary classification methods can be summarized into three categories, \ie, pre-training, prompt learning, and few-shot adaption methods, respectively.

\noindent\textbf{Pre-training Methods.}
Many pre-training efforts have been made to enhance the capabilities of VLMs in open-vocabulary classification, including large curated datasets\cite{datacomp,laion5b} and enhanced training strategies~\cite{wei2022mvp, FLIP, li2022fine, sun2023eva}. They need to retrain the model from scratch which consumes considerable time, samples, and annotations.

\noindent\textbf{Prompt Learning Methods.}
To efficiently strengthen the capabilities of VLM in classification, various prompt learning methods have been proposed. CoOp~\cite{coop} learns static contextual tokens from a few-shot dataset but tends to overfit to the training classes, degrading performance on unseen classes. To mitigate this, CoCoOp~\cite{cocoop} acquires dynamic instance-specific tokens from the input image, aiming to improve the classification of unseen classes. MaPLe~\cite{khattak2023maple} endeavors to learn multi-modal prompt tokens across different layers for both vision and language branches. 
These methods require fine-tuning on each downstream dataset, tending to overfit seen classes and lacking the generalization capability as the one in VLMs.

\noindent\textbf{Few-shot Adaptation Methods.}
The few-shot classification consists of a training phase where a model is learned on a relatively large dataset and an adaptation phase where the learned model is adapted to previously unseen tasks with limited labeled samples. Under this framework, methods can be roughly divided into two groups: meta-learning methods and non-meta-learning methods. 
A recent work~\cite{luo2023few-shot} reveals that the training and adaptation phases in few-shot image classification are completely disentangled. Besides, it also demonstrates the visual backbone pretrained with CLIP's training algorithm has superior performance than previous few-shot training algorithms. 
In our work, we take the pre-trained visual backbone of CLIP as the base model and evaluate different adaptation methods on top of it. The adaptation methods encompass
the ones from training-free methods including MatchingNet~\cite{vinyals2016matching}, Nearest Centroid Classifier(PN)~\cite{snell2017prototypical}, and the ones from training-based methods
including MAML~\cite{finn2017model}, Logistic Regression~\cite{tian2020rethinking}, Cosine Classifier~\cite{chen2019closer}, URL~\cite{li2021universal}, and CEPA~\cite{Hao_2023_ICCV}.

\subsection{Open-Vocabulary Detection}
Much recent work aims to transfer the open-vocabulary capabilities of VLMs to object detection~\cite{Zareian_2021_CVPR,ViLD,zhong2021regionclip}.
Techniques including knowledge distillation~\cite{ViLD} and prompt optimization~\cite{DetPro, wu2023cora} have been used to train an open-vocabulary detector with the pre-trained VLMs. Weak-labeling and Pseudo-boxes methods~\cite{zhong2021regionclip, GLIP,zhou2022detecting, pb-ovd,feng2022promptdet} have also been proposed to enhance the object-level recognition ability of VLMs. In addition, some works add new detection heads on the top of the pre-trained visual backbone of VLMs or SAM~\cite{han2023boosting}, either by keeping the backbone frozen~\cite{kuo2022fvlm} or finetunable~\cite{minderer2022simple,rovit}. 
Recently, pre-training the vision-language models for open-vocabulary detection is a new direction. GLIP~\cite{GLIP, zhang2022glipv2} and DetCLIP~\cite{detclip, yao2023detclipv2} train on a combination of a detection, grounding, and caption data to learn the word-region alignment. RO-ViT~\cite{rovit} proposes pretraining region-aware positional embeddings to enhance VLM's capability in dense prediction tasks. 

In addition, recent MM-OVOD~\cite{kaul2023multi} and MQ-Det~\cite{xu2023multi} introduce exemplar images to enhance the text classifier for open-vocabulary detection. However, MM-OVOD takes two modalities equally and directly calculates the arithmetic mean of the newly learned vision-based classifier and the existing textual classifier in VLMs to obtain the multimodal classifier. MQ-Det uses textual features as queries to extract information from exemplar images and refine the original text classifier with cross-attention mechanisms. This is based on the assumption that the textual modality is more important. However, influenced by the quality of exemplars and text, the preference for the two modalities should be dynamic across different categories.

\subsection{Differences with Previous Works}
Our OVMR method presents several differences with previous open vocabulary classification and detection methods.
First, unlike traditional pre-training methods, which require considerable resources, OVMR involves a lightweight visual token generator pre-trained on a smaller dataset. This enables an efficient integration of new category cues into the model without the need for fully retraining it.
Second, our approach effectively circumvents the overfitting issues inherent in prompt learning methods, as it does not learn class-specific parameters. Additionally, the plug-and-play property allows it to transfer seamlessly to various tasks after pre-training.
Third, OVMR utilizes the strong generalization capabilities of language models to adaptively fuse multi-modal cues. In contrast to methods like MM-OVOD and MQ-Det, which treat modalities equally or prioritize text, OVMR further dynamically integrates uni-modality classifiers and the multi-modal classifier by evaluating their performance. This two-stage classifier generation pipeline is more robust to scenarios with low-quality exemplars or textual descriptions, making OVMR perform substantially better in various tasks. 



%% file: sec/3_method.tex
\section{Methodology}
\label{sec:method}
\begin{figure}[t]
\centering
\includegraphics[width=1\linewidth]{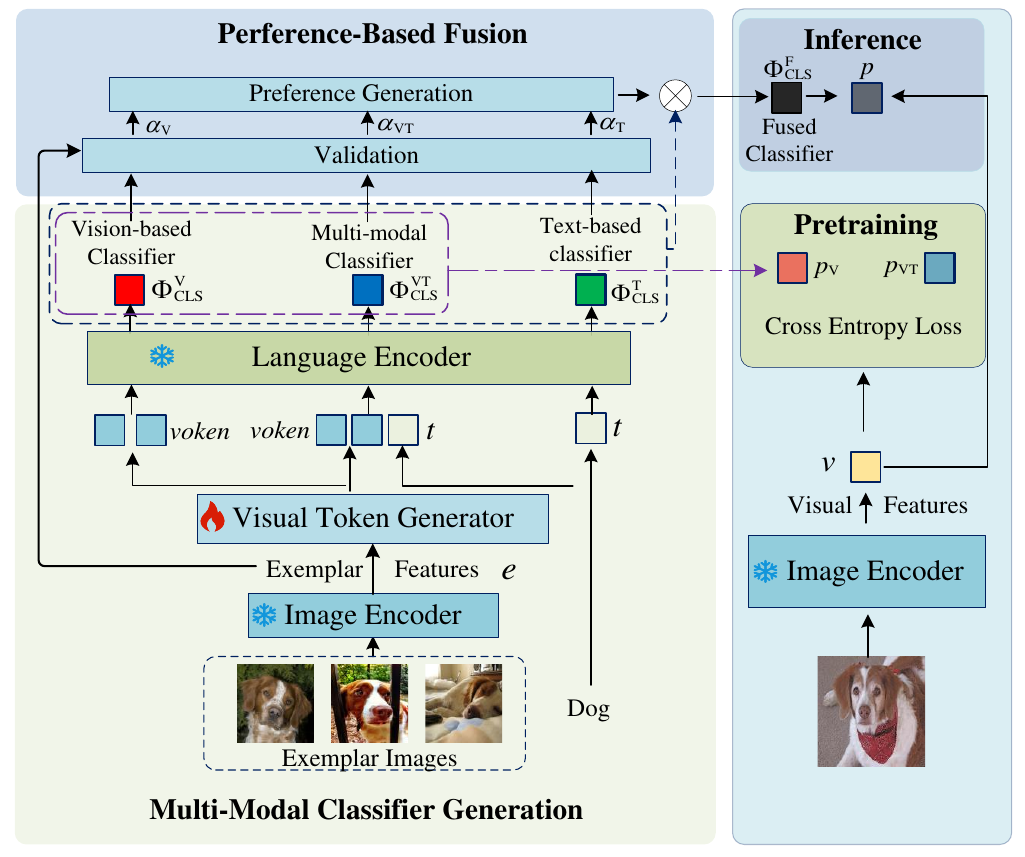}
\caption{Illustration of the pipeline for novel classifier generation (left) and image classification (right). }
\label{fig2:Architecture}
\vspace{-4mm}
\end{figure}

As illustrated in~\cref{fig2:Architecture}, our OVMR is composed of two principal modules. The first is a multi-modal classifier generation module. This module leverages a generalizable language encoder for dynamically integrating text and visual exemplars. It also includes a newly pre-trained visual token generator that embeds the exemplar images into the language space. The second module is a test-time preference-based fusion module, which does not introduce any trainable parameters. 
We will introduce the multi-modal classifier generation and the pre-training of the visual token generator in \cref{method:multi-cls-generation} and the test-time preference-based fusion in \cref{method:preference}. \cref{method:adaption_ovd} discusses the adaption of OVMR to the open-vocabulary detection task. 

\subsection{Multi-modal Classifier Generation}
\label{method:multi-cls-generation}
The multi-modal classifier generation module aims to generate a multi-modal classifier by adaptively fusing visual exemplars and textual descriptions. For a new category of interest $C_i$, we denote its visual exemplars, target images, and textual tokens as $E_i\in \mathbb{R}^{M\times H \times W \times 3}$, $V_i\in \mathbb{R}^{N\times H \times W \times 3}$ and $t_i\in \mathbb{R}^{L_i\times d}$, respectively, where $M$ is the number of exemplar images, $N$ is the number of target images, $L_i$ is the length of textual tokens belonging to $C_i$ category and $d$ is the hidden dimension of token embeddings. Furthermore, the visual exemplars $E_i$ and target images $V_i$ are encoded into exemplar features $e_i$ and visual features $v_i$ by the CLIP image encoder $\Phi_\textsc{CLIP-V}$.

In order to utilize the language encoder to generate a good multi-modal classifier, a key prerequisite is to extract robust visual tokens that the language encoder can understand from visual exemplars. These visual tokens need to accurately represent the class-discriminative visual details. 
The visual token generator consists of $P$ class-agnostic learnable queries $q \in \mathbb{R}^{P\times d}$ and four layers of transformer blocks with global self-attention. Leveraging the self-attention interaction between the learnable queries and exemplar features, the learnable queries $q$ adaptively extract visual tokens of the $C_i$ category from exemplar features $e_i$. This process can be formulated as:
 \begin{equation}
    \text{voken}_i=\Phi_\textsc{VOK} ([q, e_i]),
\end{equation}
where $[\cdot]$ represents word-level concatenation, $\text{voken}_i\in \mathbb{R}^{P\times d}$ is the corresponding output of query tokens $q$.

Then, we use the language encoder to analyze the relationships between visual and textual tokens and adaptively generate the weight $w_\textsc{VT}^i$ of the multi-modal classifier for the $C_i$ category as:
\begin{equation}
    \label{eq:w_gen}
    w_\textsc{VT}^i=\Phi_\textsc{CLIP-T}([\text{voken}_{i}, t_i]).
\end{equation}

The multi-modal classifier $\Phi_\textsc{CLS}^\textsc{VT}$ thus can be represented as:
\begin{equation}
     \label{eq:cls_gen}
    \Phi_\textsc{CLS}^\textsc{VT}(v) = \frac{\text{exp}(\text{sim}(v,w_\textsc{VT})/\tau_t)}{\sum_{k=1}^{C}\text{exp}(\text{sim}(v, w_\textsc{VT}^{k})/\tau_t)}, 
\end{equation}
where $\text{sim}(\cdot, \cdot)$ denotes cosine similarity and $\tau_t$ is the temperature.

\noindent\textbf{Pre-training of the Visual Token Generator.}
The visual token generator is the only trainable component in our OVMR, which extract class-discriminative visual information from exemplar images and affects the performance of the multimodal classifier. Thus it's important to make sure that the visual tokens can encompass the class-discriminative visual details as much as possible. We separately input the visual tokens into the language encoder to generate the vision-based classifier $\Phi_\textsc{CLS}^\textsc{V}$, and then optimize the vision-based classifier with the multi-modal classifier together.
The prediction probability of target images $v$ over the vision-based and multi-modal classifiers can be computed as:
\begin{equation}
\label{equation:logits}
    p_\textsc{V} = \Phi_\textsc{CLS}^\textsc{V}(v),\  \ \ \ p_\textsc{VT} = \Phi_\textsc{CLS}^\textsc{VT}(v).
\end{equation}

The overall pre-training objective of the visual token generator can be represented by:
\begin{equation}
\mathcal{L} = \textsc{CE}(p_\textsc{V}, \mathcal{T})+\textsc{CE}(p_\textsc{VT}, \mathcal{T}),
\end{equation}
where $\textsc{CE}(\cdot)$ denotes the cross-entropy loss, $\mathcal{T}$ are the ground-truth labels of target images.

To ensure the generalizability of the visual token generator, we design an effective pre-training strategy. During each training iteration, We randomly sample K-shot images for each category from the ImageNet21k-OVR, a subset of ImageNet21k detailed in \cref{exp:datasets}. From these images, $M$ images are randomly selected as visual exemplars for each category, while the remaining $N$=$K$-$M$ images are used as target images. The number of exemplars $M$ is varied randomly within the range $[K/4, 3K/4]$ to simulate diverse scenarios encountered in practical applications.
It is important to ensure that there is no overlap between exemplar and target images, which guides the model to learn class-discriminative features rather than instance-specific details. Additionally, we have incorporated techniques such as random path dropout in the attention layer and channel-wise dropout in each transformer block of the visual token generator to further enhance its generalizability.

\subsection{Preference-Based Fusion}
\label{method:preference}

The preference-based fusion aims to simulate the test-time validation results on exemplar images to measure the preference of text-based, vision-based, and multi-modal classifiers. It hence generates a more powerful fused classifier based on the estimated preference. And the text-based classifier $\Phi_\textsc{CLS}^\textsc{T}$ can be acquired conveniently by feeding textual tokens into a language encoder. 

The preference-based fusion process begins by validating different classifiers on exemplar images. Similar to \cref{equation:logits}, by inputting the exemplar features $e$ into various classifiers, we obtain the prediction probabilities of exemplar images over different classifiers. These probabilities, $\hat{p}_\textsc{VT}$, $\hat{p}_\textsc{V}$, and $\hat{p}_\textsc{T}$, correspond to the multi-modal, vision-based, and text-based classifiers, respectively.
Then, based on these prediction probabilities, we can get each category's preferences for different classifiers:
\begin{equation}
    \begin{aligned}
        &\alpha_\textsc{T}=\Omega(\hat{p}_\textsc{T}, \mathcal{T}_E), \\
        &\alpha_\textsc{V}=\Omega(\hat{p}_\textsc{V}, \mathcal{T}_E), \\
        &\alpha_\textsc{VT}=\Omega(\hat{p}_\textsc{VT}, \mathcal{T}_E),
    \end{aligned}
\end{equation}
where $\alpha_\textsc{T},\alpha_\textsc{V},\alpha_\textsc{VT}\in \mathbb{R}^{|C|\times 1}$ denotes the preferences for text-based, vision-based, and multi-modal classifiers, separately. $\mathcal{T}_E$ are the ground-truth labels of exemplar images. $\Omega$ denotes the chosen evaluation metric, which is the F1 score~\cite{sokolova2009systematic} in open-vocabulary classification tasks. We select the F1 score because it stably reflects the quality of a classifier by comprehensively considering both precision and recall.
The preferences for different classifiers can be denoted as:
\begin{equation}
    [\hat{\alpha}_\textsc{V}, \hat{\alpha}_\textsc{VT}, \hat{\alpha}_\textsc{T}] = \sigma(\tau_p[\alpha_\textsc{V}, \alpha_\textsc{VT},  \alpha_\textsc{T}]),
\end{equation}
where $\sigma$ denotes softmax function, $\tau_p$ is the temperature for preference generation, and $[\cdot]$ represent concatenation at the last dimension. 

Then the fused classifier of the multi-modal classifier and uni-modal classifiers can be formulated as:
\begin{equation}
    \Phi_\textsc{CLS}^F = 
    \hat{\alpha}_\textsc{V}\cdot \Phi_\textsc{CLS}^\textsc{V}+
    \hat{\alpha}_\textsc{VT}\cdot \Phi_\textsc{CLS}^\textsc{VT}+
    \hat{\alpha}_\textsc{T} \cdot \Phi_\textsc{CLS}^\textsc{T}.
\end{equation}
We hence can get the final prediction probability $p$ of the target images over the fused classifier as follows:
\begin{equation}
    p = \Phi_\textsc{CLS}^F(v).
\end{equation}
The preference-based fusion effectively leverages exemplar images to boost the recognition ability of VLMs, without introducing any trainable parameters. 

\subsection{Adaptation to Open-Vocabulary Detection}
\label{method:adaption_ovd}
Our method is not limited to the open-vocabulary classification. It can be flexibly adapted to other open vocabulary recognition tasks like detection. In this work, we make use of a popular multi-stage detector based on
CenterNet2~\cite{zhou2021probabilistic} as done in Detic~\cite{zhou2022detecting} and MM-OVOD~\cite{kaul2023multi}.
For simplicity, we consider the two-stage variant of this classifier. Using $\left\{{b}_j, p^j\right\}^{H}_{j=1}$ to denote its output, and $H$ as the number of proposals, it can be formulated as:
\begin{align}
&\left\{r_j\right\}^{H}_{j=1} = \label{eq:a}
\Phi_\textsc{roi}\circ\Phi_\textsc{pg}\circ
\Phi_\textsc{enc}\left({I}\right),  \\
&\left\{
{b}_j, p^j \right\}^{H}_{j=1} = \label{eq:b}
\left\{
\Phi_\textsc{bbox}\left(r_j\right), 
\Phi_\textsc{CLS}\circ\Phi_\textsc{proj}\left(r_j\right)
\right\}^{H}_{j=1} ,
\vspace{-20pt}
\end{align}
where each input image $I$ is first sequentially processed by a set of operations:
a trainable image encoder~$\left(\Phi_\textsc{enc}\right)$,
a proposal generator~$\left(\Phi_\textsc{pg}\right)$,
a region-of-interest (RoI) feature pooling module~$\left(\Phi_\textsc{roi}\right)$,
finally yielding a set of RoI features $\left\{r_j\right\}^{H}_{j=1}$.
The RoI features are processed by a bounding box module~$\left(\Phi_\textsc{bbox}\right)$ to infer position of objects, 
$\left\{{b}_j\right\}^{H}_{j=1}$.

Additionally, the RoI features are processed by a classification module,
consisting of a linear projection~$\left(\Phi_\textsc{proj}\right)$,
and multi-modal classifiers~$\left(\Phi_\textsc{CLS}\in\{\Phi_\textsc{CLS}^{\text{VT}}, \Phi_\textsc{CLS}^{\text{V}}, \Phi_\textsc{CLS}^{\text{T}}\}\right)$,
yielding prediction probabilities of RoI features, $\left\{p^j\right\}^{H}_{j=1}$, where $p^j\in\{p^j_\textsc{VT}, p^j_\textsc{V}, p^j_\textsc{T}\}$.

During training, the multi-modal, vision-based, and text-based classifiers are kept frozen while other components in the detector are trainable. We incorporate a sigmoid cross entropy loss for each classifier.
In preference-based fusion, the recently trained detector, along with the multi-modal classifier, is utilized to identify the most accurate pseudo box within the ground-truth label for each exemplar image. Subsequently, with these pseudo annotations, we calculate the Average Precision (AP) for each class, which aids in computing the preference for different classifiers.  

%% file: sec/4_experiments.tex
\section{Experiments}
\label{sec:exp}
\subsection{Datasets}
\label{exp:datasets}
\noindent\textbf{Classification:}
Following prompt learning methods~\cite{cocoop, khattak2023maple}, we use the 11 image classification datasets, which cover a diverse set of recognition tasks. Specifically, the benchmark includes ImageNet~\cite{deng2009imagenet} and Caltech101~\cite{fei2004learning} for classification on generic objects; OxfordPets~\cite{parkhi2012cats}, StanfordCars(Cars)~\cite{krause20133d}, Flowers102~\cite{nilsback2008automated}, Food101~\cite{bossard2014food} and FGVCAircraft(Aircraft)~\cite{maji2013fine} for fine-grained classification; SUN397~\cite{xiao2010sun} for scene recognition; UCF101~\cite{soomro2012ucf101} for action recognition; DTD~\cite{cimpoi2014describing} for texture classification; and finally EuroSAT~\cite{helber2019eurosat} for satellite imagery recognition. 

\noindent\textbf{Detection:} 
In the open-vocabulary detection, the LVIS~\cite{gupta2019lvis} dataset is used following previous works~\cite{ViLD,zhou2022detecting}, which contains 100K images with 1,203 classes. The classes are divided into three groups, namely frequent, common and rare, based on the number of training images. We treat 337 rare classes as novel classes and use the frequent and common classes as base classes for training. When using image classification data as extra weak supervision, we use the subset of categories in ImageNet21k that overlap with the LVIS vocabulary and denote this subset as IN-L, as in Detic~\cite{zhou2022detecting}.

\noindent\textbf{Pre-training Set:}
We construct the pre-training dataset based on ImageNet21k. To prevent data leakage, we remove any overlapping categories present in both the 11 classification datasets and the LVIS dataset from ImageNet21k. Additionally, we limit the number of images per category to 64 to improve pre-training efficiency. As a result, we create a 64-shot subset of ImageNet21k, termed ImageNet21k-OVR. This subset contains 18,631 categories and encompasses a total of 1.1 million images, 
which is much smaller than the dataset used for the VLM's pre-training.

\begin{table*}[t]
\centering
\footnotesize
\setlength{\tabcolsep}{2.5pt}
\caption{Open-Vocabulary Classification Results in Prompt Learning Setup. Our method achieves comparable performance to the SoTA method using the same 16-shot exemplar images without any extra training. $\dagger$: integrating CoOP's prompt tokens into our method.}
\begin{tabular}{c|c|ccccccccccc|c}
\toprule
    Methods & Extra Training & ImageNet  & Caltech101 & OxfordPets & Cars & Flowers102 & Food101 & Aircraft & SUN397 & DTD & EuroSAT & UCF101 & Average \\
    \midrule
CLIP~\cite{CLIP}  & $\times$ & 72.43 & 96.84 & 91.17 & 63.37 & 72.08 & 90.10 & 27.19 & 69.36 & 53.24 & 56.48 & 70.53 & 69.34 \\
CoOp~\cite{coop}  & $\checkmark$ & 76.47 & 98.00 & 93.67 & 78.12 & 97.60 & 88.33 & 40.44 & 80.60 & 79.44  & 92.19 & 84.69 & 82.69 \\
CoCoOp~\cite{cocoop}\ & $\checkmark$ & 75.98 & 97.96 & 95.20 & 70.49 & 94.87 & 90.70 & 33.41 & 79.74 & 77.01 & 87.49 & 82.33 & 80.47 \\
MaPLe~\cite{khattak2023maple} & $\checkmark$ & 76.66 & 97.74 & \textbf{95.43} & 72.94 & 95.92 & \textbf{90.71} & 37.44 & 80.82 & 80.36 & \textbf{94.07} & 83.00 & 82.28 \\ \hline
\textbf{OVMR} & $\times$  & 76.77 & 98.00 & 94.97 & 73.93 & 97.83 & 89.93 & 40.37 & 81.83 & 77.10 & 90.00 & 85.03 & 82.34 \\
\textbf{OVMR$\dagger$} & $\checkmark$ & \textbf{76.87} & \textbf{98.27} & 94.23 & \textbf{79.67} & \textbf{98.53} & 89.13 & \textbf{43.07} & \textbf{82.40} & \textbf{80.37} & 93.07 & \textbf{85.30} & \textbf{83.72}
\\ \bottomrule
\end{tabular}
\label{tab1:prompt learning methods}
\end{table*}

\begin{table*}[t]
\centering
\footnotesize
\setlength{\tabcolsep}{2pt}
\caption{Open-Vocabulary Classification Results in Few-shot Setup. Our method shows superior performance on average.}
\begin{tabular}{c|c|ccccccccccc|c}
\toprule
Methods & Extra Training & ImageNet  & Caltech101 & OxfordPets & Cars & Flowers102 & Food101 & Aircraft & SUN397 & DTD & EuroSAT & UCF101 & Average \\
    \midrule
MAML~\cite{finn2017model}  & $\checkmark$ & 56.93 & 92.97 & 33.97 & 59.10 & 76.50 & 67.83 & 26.07 & 41.67 & 62.60 & 83.80 & 69.17 & 60.96 \\
MatchingNet~\cite{vinyals2016matching}  & $\times$ & 55.87 & 95.57 & 75.53 & 58.17 & 91.83 & 79.97 & 33.60 & 68.17 & 66.33 & 82.73 & 75.43 & 71.20 \\
LR~\cite{tian2020rethinking}  & $\checkmark$ & 63.67 & 97.00 & 84.60 & 66.07 & 94.90 & 84.70 & 37.93 & 75.37 & 74.43 & 88.40 & 80.57 & 77.06 \\
PN~\cite{snell2017prototypical}   & $\times$ & 65.77 & 96.13 & 83.37 & 76.30 & 96.13 & 84.93 & 40.43 & 76.30 & 72.20 & 86.87 & 80.53 & 77.32\\ 
CEPA~\cite{Hao_2023_ICCV}  & $\checkmark$ & 69.00 & 97.10 & 88.10 & 71.70 & 96.10 & 85.20 & 39.10 & 78.10 & 73.70 & 90.70 & 80.40 & 79.02  \\ 
CC~\cite{chen2019closer}   & $\checkmark$ & 70.83 & 97.23 & 86.53 & 75.27 & 96.97 & 87.07 & 41.30 & 79.53 & 76.83 & 91.07 & 83.00 & 80.51 \\
URL~\cite{li2021universal} & $\checkmark$ & 72.07 & 97.77 & 89.27 & \textbf{78.17} & 97.23 & 87.40 & \textbf{44.53} & 79.97 & \textbf{80.77} & \textbf{92.00} & 82.13 & 82.13 \\ \hline

\textbf{OVMR} & $\times$  & \textbf{76.77} & \textbf{98.00} & \textbf{94.97} & 73.93 & \textbf{97.83} & \textbf{89.93} & 40.37 & \textbf{81.83} & 77.10 & 90.00 & \textbf{85.03} & \textbf{82.34} \\
\bottomrule
\end{tabular}
\vspace{-5mm}
\label{tab2:few-shot-methods}
\end{table*}

\subsection{Implementation Details}
\noindent\textbf{Open-Vocabulary Classification.}
Following prompt learning methods~\cite{cocoop, khattak2023maple}, we select ViT-B/16 of CLIP as our base model and pre-train a plug-and-play visual token generator to enhance its recognition capability. The sample number $K$ of each class is set to 8 and we sample 192 classes per batch, which results in a total batch size of 1536. We pre-train the visual token generator on ImageNet21k-OVR for 30 epochs in 12 hours on a single 3090 GPU. We adopt an Adam optimizer and a cosine learning rate scheduler, where the learning rate is set to 0.0002. The number $P$ of visual tokens is set to 2. The $\tau_p$ in preference-based fusion is set to 10. 

\noindent\textbf{Open-Vocabulary Detection.} 
The architecture and training recipe is almost identical to that in Detic~\cite{zhou2022detecting} and MM-OVOD~\cite{kaul2023multi}, using the CenterNet2 model with a ResNet-50 or Swin-B backbone pre-trained on ImageNet21k-P. 
It's worth noting that we re-implement a memory-efficient version of Detic by limiting the maximum number of ground-truth boxes per image to 10 following~\cite{zhou2019objects, wang2023contextual} in order to reproduce Detic's experiments on four 24GB 3090 GPUs. Based on the memory-efficient version, we replace the original classifier with our proposed multi-modal classifiers and further introduce the preference-based fusion. Besides, for a fair comparison, following previous works~\cite{zhou2022detecting}, we pre-train a new visual token generator for ViT-B/32 of CLIP on ImageNet21k-OVR with LVIS base categories. 
In system-level implementation, we take Swin-B as our backbone and train the detector with additional image-labeled data(IN-L) following Detic~\cite{zhou2022detecting}.

\noindent\textbf{Training and Test Setups:}
In open-vocabulary classification, we select the first half of the categories in the classification dataset as base classes and conduct comparisons on base classes for a fair comparison with existing methods. In the prompt learning setup, during inference, we utilize the same 16-shot training images used in previous methods to serve as exemplar images. Prompt learning methods need to fine-tune for each downstream dataset using these 16-shot images, while our OVMR does not require any extra training. The test set is also kept the same as prompt learning methods.
In the traditional few-shot setup, following the recent work~\cite{luo2023few-shot}, we take the visual encoder of CLIP as the base model and evaluate different few-shot adaptation algorithms in the same 16-shot setup as prompt learning methods. 
In open-vocabulary detection, the exemplar images are the same as those in MM-OVOD~\cite{kaul2023multi}, which provides 5-shot exemplar images for each category. 
The main evaluation metric is the mask AP averaged over the “rare” classes, which is denoted as mask AP$_r$.

\subsection{Comparison with Recent Works}

\noindent\textbf{Comparison with Prompt Learning Methods.}
In ~\cref{tab1:prompt learning methods}, we compare our method with existing prompt learning methods on 11 classification datasets. Our methods yield performance comparable to the state-of-the-art method CoOp without any extra training in downstream tasks. It's worth noting that the pre-training dataset, ImageNet21k-OVR, used for the visual token generator has been curated to exclude the categories from the 11 classification datasets and LVIS. The categories in these evaluation datasets are only seen at test time. Providing a few exemplar images during inference, our method manages to match the performance of existing prompt learning methods. 
Further, by integrating task-specific prompt tokens used in CoOp into our approach and fine-tuning these tokens under the same conditions as CoOp, we observe an enhanced performance, which surpasses CoOp by an average accuracy of 1.03$\%$ across the 11 datasets. These results demonstrate that our method can not only enhance open-vocabulary classification without additional training but also be compatible with other optimization-based prompt learning approaches.

\noindent\textbf{Comparison with Traditional Few-shot Methods.} We compare our method with few-shot methods in ~\cref{tab2:few-shot-methods}. Our training-free method exhibits superior performance over the state-of-the-art training-based method URL, both on average and across the majority of classification datasets.  Particularly in datasets encompassing a broad range of categories, our method significantly outperforms URL, as evidenced by a notable margin of 4.7$\%$ in ImageNet. Conversely, in datasets with a few classes, like EuroSAT, URL shows superior performance. This is because supervised fine-tuning tends to be more effective in easy scenarios with fewer classes. In contrast with the training-free method PN, our approach shows a substantial performance gain in complex datasets with hundreds of classes. For instance, on ImageNet, we observe an impressive improvement of 11.0$\%$.

\begin{table}[t]
\centering
\footnotesize
\renewcommand{\arraystretch}{0.9}
\setlength{\tabcolsep}{6.5pt}
\caption{LVIS Open-Vocabulary Detection. 
$\ddagger$: reporting box AP. ``OVMR$_\text{T}$'': pure text-based classifier.}
\begin{tabular}{llccc}
\toprule
Method & \makecell{Detector \white{----} \\ Backbone \white{----}}  & \makecell{Extra\\ Data} &
\makecell{ {AP$_r$}}
& \makecell{ \gray{AP}}  

\\
\midrule
\multicolumn{4}{l}{\bf\textit{ResNet50 Comparison:}} \\
ViLD-Ens~\citep{ViLD}            & RN50    & $\times$ & 16.6     & \gray{25.5}   \\
OV-DETR~\citep{zang2022open}            & RN50  & $\times$  & 17.4     & \gray{26.6}\\
Detic~\citep{zhou2022detecting}            & RN50 &   $\times$    & 17.8      & \gray{26.8} \\
F-VLM~\citep{kuo2022fvlm}            & RN50  & $\times$  & 18.6      & \gray{24.2}  \\
PromptDet~\citep{feng2022promptdet}       & RN50 &  $\times$  & 19.0      & \gray{21.4} \\
BARON~\citep{wu2023aligning}   & RN50   & $\times$ & 19.2 & \gray{26.5} \\

DetPro~\citep{du2022learning}     & RN50   &    $\times$   & 19.8      & \gray{25.9} \\

MM-OVOD$_\textbf{T}$~\citep{kaul2023multi}        & RN50   &    $\times$   & 19.3      & \gray{30.3} \\
MM-OVOD~\citep{kaul2023multi}      & RN50   &    $\times$  & 19.3      & \gray{30.6} \\

\textbf{OVMR$_\text{T}$}    & RN50     &  $\times$  & 19.0   & \gray{29.6} \\
\textbf{OVMR}   & RN50      &  $\times$ & \textbf{21.2}   & \gray{30.0}  \\

\midrule
\multicolumn{4}{l}{\bf\textit{System-level Comparison:}} \\
RegionCLIP~\citep{zhong2021regionclip}      & RN50x4\gray{(87M)}     & $\checkmark$   & 22.0      & \gray{32.3} \\
CondHead~\citep{wang2023learning}        & RN50x4\gray{(87M)} & $\times$       & 24.4      &  \gray{32.0}  \\
OWL-ViT$\ddagger$~\citep{minderer2022simple}         & ViT-L/14\gray{(303M)}   & $\checkmark$ & 25.6     & \gray{34.7}  \\
F-VLM~\citep{kuo2022fvlm}     & RN50x4\gray{(87M)} & $\times$     & 26.3      & \gray{28.5} \\
{RO-ViT~\citep{rovit}}    & ViT-B/16\gray{(86M)}   & $\checkmark$   & {28.0} & \gray{30.2} \\
CORA$\ddagger$~\citep{wu2023cora}      & RN50x4\gray{(87M)}   & $\checkmark$   &28.1 &- \\
CFM-ViT~\citep{kim2023contrastive}    & ViT-B/16\gray{(86M)}   &  $\checkmark$ & {28.8} & \gray{32.0} \\
CoDet~\citep{ma2023codet}   & Swin-B\gray{(88M)}    &  $\checkmark$ & {29.4} & \gray{39.2}  \\
{RO-ViT~\citep{rovit}}    & ViT-L/16\gray{(303M)} & $\checkmark$      & {32.1} & \gray{34.0} \\
DITO~\citep{kim2023detection}   & ViT-B/16\gray{(86M)}   & $\checkmark$   & {32.5} & \gray{34.0} \\
Detic~\citep{zhou2022detecting} & Swin-B\gray{(88M)}&$\checkmark$  & 33.8 & \gray{40.7} \\
\textbf{OVMR$_\text{T}$}    & Swin-B\gray{(88M)}    & $\checkmark$     & 33.3      & \gray{40.8} \\
\textbf{OVMR}    & Swin-B\gray{(88M)}    & $\checkmark$     & \textbf{34.4}      & \gray{40.9}\\
\bottomrule
\end{tabular}
\vspace{-5mm}
\label{tab3:ovd_lvis}
\end{table}

\noindent\textbf{Comparison with Open-Vocabulary Detection Methods.}
~\cref{tab3:ovd_lvis} presents our open-vocabulary detection results on LVIS. In the ResNet50 comparison, our OVMR, which utilizes multi-modal classifiers and preference-based fusion, outperforms the existing approach MM-OVOD using the same exemplar images by 1.9\% mask AP$_r$ on novel categories. 
In system-level comparison, follow Detic~\citep{zhou2022detecting}, OVMR achieves a superior performance with additional weak supervision from IN-L, which outperforms recent pretraining-based method DITO by 1.9\% on mask AP$_r$ on novel categories. Using only the text classifier, we achieve a mask AP$_r$ of 33.3\%, which is marginally lower by 0.5\% compared to the 33.8\% mask AP$_r$ reported by Detic. However, our multi-modal version, OVMR, reaches 34.4\% mask AP$_r$, surpassing the original Detic by 0.6\% in mask AP$_r$. Moreover, OVMR$_\text{T}$ and OVMR share the same detector parameters. When there are no exemplar images, our method can still work with only textual descriptions.

\subsection{Ablation Study}
\noindent\textbf{Effectiveness of Our Proposed Components.} 
In ~\cref{tab4:ablation on components}, we systematically evaluate each component of our proposal through ablation studies on 11 classification tasks. Our baseline method, OVMR$_\text{T}$, employs only a text-based classifier. In contrast, OVMR$_\text{V}$ utilizes solely a vision-based classifier, while OVMR$_\text{VT}$ integrates a multi-modal classifier. A comparative analysis of OVMR$_\text{VT}$ against OVMR$_\text{T}$ and OVMR$_\text{V}$ reveals the superior performance of the multi-modal classifier. 
OVMR$_\text{VT}^-$ takes $K$ sampled images both as exemplars and target images and removes the dropout operations. The performance disparity between OVMR$_\text{VT}$ and OVMR$_\text{VT}^-$ underscores the effectiveness of our pre-training strategy depicted in \cref{method:multi-cls-generation}.
As demonstrated in OVMR$_{{\text{T+V, T+VT, V+VT}}}$ and OVMR, the preference-based fusion of different classifiers further elevates performance, achieving a best average accuracy of 82.34\% across 11 datasets when fusing all three classifiers. This marks a significant 15.54\% improvement over the baseline text-based classifier. 
Parallel ablation studies and consistent results for our components in open-vocabulary detection are presented in ~\cref{tab5:ablation on components ovd}.

\begin{table}[t]
    \centering
    \footnotesize
    \renewcommand{\arraystretch}{0.9}
    \setlength{\tabcolsep}{6.5pt}
    \caption{Ablation study of each component in open-vocabulary classification. ``T'': text-based classifier. ``V'': vision-based classifier. ``PS'': pre-training strategy of the visual token generator. ``VT'': multi-modal classifier. ``Fusion'': preference-based fusion.} 
    \begin{tabular}{c|ccccc|c}
    \toprule
    Methods  & T & V &\makecell{PS}& \makecell{VT} & \makecell{Fusion} & \makecell{Average} \\ \midrule
    OVMR$_\text{T}$   &   $\checkmark$   &        &   &          &        &   66.80        \\
    OVMR$_\text{V}$   &      &   $\checkmark$     &   $\checkmark$  &        &        &   80.28       \\
    OVMR$_\text{VT}^-$  &      &      &  &   $\checkmark$ &        &     77.86     \\
    OVMR$_\text{VT}$  &     &     & $\checkmark$ &       $\checkmark$      &        &     80.99     \\
    OVMR$_\text{T+V}$  &  $\checkmark$   &  $\checkmark$   &  &          &     $\checkmark$   &     81.71     \\
    OVMR$_\text{T+VT}$  & $\checkmark$    &    & $\checkmark$ &       $\checkmark$      &    $\checkmark$    &     81.30     \\
    OVMR$_\text{V+VT}$  &     &  $\checkmark$   & $\checkmark$ &       $\checkmark$      &   $\checkmark$     &     81.96     \\
    OVMR  &   $\checkmark$   &    $\checkmark$  & $\checkmark$ &      $\checkmark$       &     $\checkmark$   &     \textbf{82.34}     \\ \bottomrule
    \end{tabular}
    \label{tab4:ablation on components}
\end{table}

\begin{table}[t]
    \centering
    \footnotesize
    \renewcommand{\arraystretch}{0.9}
    \setlength{\tabcolsep}{6.5pt}
    \caption{Ablation study of each component in open-vocabulary detection. AP$_r$ is reported.} 
    \begin{tabular}{c|cccc}
    \toprule
    Backbone & OVMR$_\text{T}$ & OVMR$_\text{V}$ & OVMR$_\text{VT}$& OVMR \\
    \midrule
    RN50   &   19.0   &   15.9     &  21.0 &  \textbf{21.2}           \\
    Swin-B   &   33.3   &   31.9    &   33.8  & \textbf{34.4}  \\ \bottomrule
    \end{tabular}
    \label{tab5:ablation on components ovd}
    \vspace{-5mm}
\end{table}

\noindent\textbf{Superiority of Our Proposed Fusion.}
Our fusion method is compared against static fusion with arithmetic mean and text-based dynamic fusion, as detailed in ~\cref{tab6:fusion_dynamic}. 
Our dynamic fusion method outperforms the static fusion method MM-OVOD, achieving a 2.4\% improvement despite MM-OVOD's better vision-based classifier performance.
Text-guided dynamic fusion method MQ-Det performs the worst, as the quality of the text query may be unreliable and this fusion largely disrupts the original features of CLIP. The results in~\cref{tab6:fusion_dynamic} demonstrate the superiority of our fusion method. 
In~\cref{tab7:fusion_ovd}, we further compare our fusion method with MM-OVOD and MQ-Det in open-vocabulary detection. When using the same 5-shot exemplar images. Our OVMR outperforms these methods in mask AP$_r$ by a notable margin.

\begin{table}[t]
    \centering
    \footnotesize
    \renewcommand{\arraystretch}{0.9}
    \setlength{\tabcolsep}{6pt}
    \caption{Comparative analysis of fusion methods in classification. ``CLIP-F'': Forming a vision-based classifier by averaging CLIP's visual features of exemplar images and statically averaging it with text-based classifier.
``MM-OVOD'': Pre-training a discriminative vision-based classifier and statically averaging it with text-based classifier.
``MQ-Det'': Utilizing the text feature as a query to dynamically aggregate related visual details from exemplar images.}
    \begin{tabular}{c|c|cccc}
    \toprule
    \makecell{Fusion\\Methods} & Dynamic  & Text  & Vision    & \makecell{Multi-Modal}    \\ \midrule
    CLIP-F   &  $\times$  & 66.80 & 77.32      & 79.92          \\
    MM-OVOD~\cite{kaul2023multi}  & $\times$   & 66.80 & 80.44      & 79.94          \\
    MQ-Det~\cite{xu2023multi}  & $\checkmark$ & 66.80 & -        & 62.60          \\
    OVMR      & $\checkmark$  & 66.80 & 80.28 & \textbf{82.34} \\ 
    \bottomrule
    \end{tabular}
    \label{tab6:fusion_dynamic}
\end{table}

\begin{table}[t]
    \centering
    \footnotesize
    \renewcommand{\arraystretch}{0.9}
    \setlength{\tabcolsep}{7pt}
    \caption{Comparison with other methods using exemplar images in open-vocabulary detection.}
    \begin{tabular}{c|c|c|cc}
    \toprule
    \makecell{Fusion\\Methods} & Dynamic & Backbone  &  AP$_r$ & \gray{mAP} \\
    \midrule
    MM-OVOD~\cite{kaul2023multi} & $\times$ & ResNet-50     &  19.3  &  \gray{30.6}         \\
    MQ-Det~\cite{xu2023multi} & $\checkmark$ & Swin-T &     15.4   &  \gray{22.6}        \\
    OVMR  & $\checkmark$  & ResNet-50  & \textbf{21.2}  &   \gray{30.0}       \\
    \bottomrule
    \end{tabular}
    \label{tab7:fusion_ovd}
\end{table}

\noindent\textbf{Evaluation Metric in Preference Generation.}
In \cref{tab8:ablation_compute_method}, directly averaging the predictions of three classifiers with static ``Mean'' can improve the accuracy to 81.45\%, which shows that uni-modal classifiers contain some distinct features that are beneficial for classification.
Besides, the accuracy of the F1 metric is superior to other evaluation metrics, because recall and precision cannot comprehensively evaluate the quality of a classifier. 

\begin{table}[t]
    \centering
    \footnotesize
    \renewcommand{\arraystretch}{0.9}
    \setlength{\tabcolsep}{10pt}
    \caption{Comparison between different evaluation metrics in open-vocabulary classification. ``Mean'': treating three classifiers as equally important, assigning them identical preferences.}
    \begin{tabular}{c|cccc}
    \toprule
    OVMR$_\text{VT}$ & Mean & Precision & Recall & F1  \\
    \midrule
    80.99 & 81.45 & 82.03 & 81.70 & \textbf{82.34} \\
    \bottomrule
    \end{tabular}
    \vspace{-5mm}
    \label{tab8:ablation_compute_method}
\end{table}

\subsection{Analysis of Exemplar Images}
\begin{figure}
    \centering
    \includegraphics[width=1\linewidth]{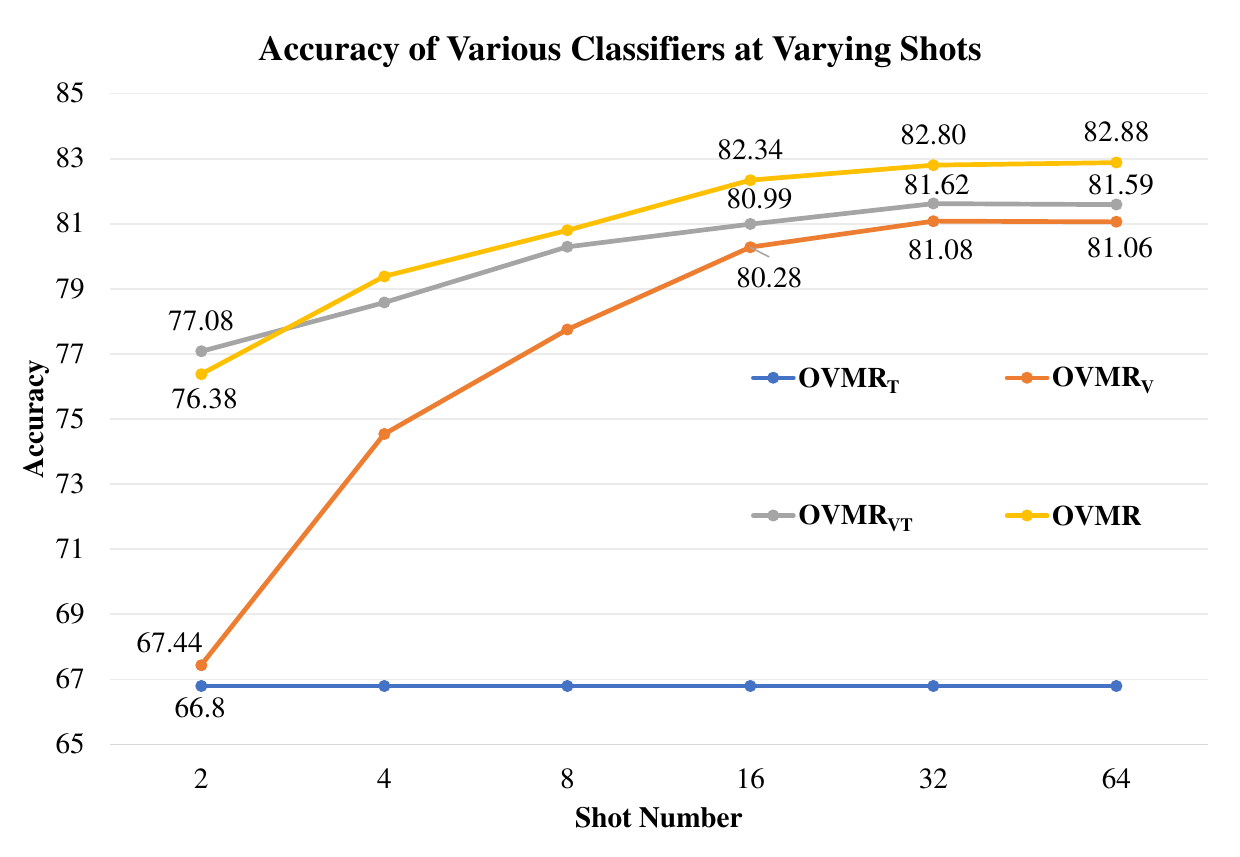}
    \caption{The accuracy of various classifiers at varying shots.}
    \label{fig3:shot_num}
\end{figure}

\noindent\textbf{The Number of Exemplar Images.}
As shown in \cref{fig3:shot_num}, with increasing the shot number of exemplar images in open-vocabulary classification, the average performance of classifiers using visual clues steadily improves. However, when it comes to 16 or 32 shots, the performance of each classifier with visual clues tends to saturate, and there is almost no improvement when increased to 64 shots. It indicates that for most classification tasks, 16-shot exemplars are enough to efficiently boost the recognition capability. In addition, in the 2-shot setting, even though the performance of the vision-based classifier is close to that of the text-based classifier, a significant improvement can be achieved using our multi-modal classifier generation module, which further demonstrates the effectiveness of our method. However, there is a small drop after introducing the preference-based fusion in the 2-shot setting. It is an inherent limitation of preference-based fusion, as fewer images may not robustly evaluate each category's preference for various classifiers. But considering that it is easy to obtain multiple exemplars of one category from the Internet, we can alleviate the negative influence by collecting more exemplars.

\noindent\textbf{Sources of Exemplar Images.}
We also evaluate our OVMR on ImageNet in a 16-shot setting with the exemplar images crawled from the Internet. The results in \cref{tab9:ablation_exemplar_source} indicate that leveraging the uncurated, readily available exemplar images from the Internet can also significantly enhance open-vocabulary recognition by a notable improvement.

\begin{table}[t]
    \centering
    \footnotesize
    \renewcommand{\arraystretch}{0.9}
    \setlength{\tabcolsep}{7pt}
    \caption{Evaluate our method on ImageNet with exemplar images from different sources. OVMR$_\text{T}$ is the zero-shot accuracy of CLIP.}
    \begin{tabular}{c|cccc}
    \toprule
    Source & OVMR$_\text{T}$ & OVMR$_\text{V}$ & OVMR$_\text{VT}$ & OVMR  \\
    \midrule
    Internet & 66.80  & 69.83 & 73.97 & \textbf{74.56} \\
    ImageNet & 66.80 & 72.63 & 76.50 &  \textbf{76.77} \\
    \bottomrule
    \end{tabular}
    \vspace{-6mm}
    \label{tab9:ablation_exemplar_source}
\end{table}

%% file: sec/5_conclusion.tex
\section{Conclusion}
In this paper, we propose a plug-and-play method, OVMR, to embed multi-modal clues of novel classes into VLMs to enhance their capability in open-vocabulary recognition.
It initially utilizes the multi-modal classifier generation module to embed the exemplar images into visual tokens and then adaptively fuse multi-modal cues by inferring their contextual relationships with the language encoder.  To alleviate the negative effects of the low-quality modality, we further propose a parameter-free fusion module to dynamically integrate the multi-modal classifier with two uni-modal classifiers by each category's specific preference for these classifiers. Extensive experiments validate our method's superior performance in both open-vocabulary classification and detection tasks, underscoring its potential to advance open-vocabulary recognition.

%% file: sec/X_suppl.tex
\clearpage
\setcounter{page}{1}
\maketitlesupplementary
\appendix
\section{Ablation Study}
\subsection{The Number of Visual Tokens}
\cref{tab:ablation_token_num} demonstrates the impact of different numbers of visual tokens on the multi-modal and fused classifiers. We observe the highest average performance for both classifiers across 11 classification datasets when P is set to 2. Consequently, the number of visual tokens is set to 2.

\begin{table}[!h]
    \centering
    \footnotesize
    \caption{Impact of different numbers of visual tokens on average accuracy across 11 classification datasets.}
    \renewcommand{\arraystretch}{0.9}
    \setlength{\tabcolsep}{10pt}
    \begin{tabular}{c|ccc}
    \toprule
    $P$ & OVMR$_\text{VT}$ & OVMR \\
    \midrule
    1 & 80.69 & 82.16 \\
    2& \textbf{80.99} & \textbf{82.34} \\
    4& 80.95 & 82.23 \\
    8& 80.54 & 82.17 \\
    \bottomrule
    \end{tabular}
    \label{tab:ablation_token_num}
\end{table}

\subsection{Temperature in Preference Generation}
\cref{tab:ablation_tau} illustrates that setting $\tau_p$ to 10 yields the best average performance for the fused classifier. Furthermore, the performance variations under different $\tau_p$ values are not significant, suggesting that our method is insensitive to $\tau_p$. This observation validates the robustness of our proposed preference-based fusion approach. Based on these findings, we set $\tau_p$ to 10 in our experiments.

\begin{table}[!h]
    \centering
    \footnotesize
    \caption{Average accuracy of the fused classifier with different $\tau_p$ across 11 classification datasets.}
    \renewcommand{\arraystretch}{0.9}
    \setlength{\tabcolsep}{10pt}
    \begin{tabular}{c|cccc}
    \toprule
    $\tau_p$ &  OVMR \\
    \midrule
    1 & 81.82\\
    10&  \textbf{82.34}\\
    20& 82.07 \\
    \bottomrule
    \end{tabular}
    \label{tab:ablation_tau}
\end{table}

\section{Comparison on Novel Sets}
By embedding multi-modal clues of novel categories into vision-language models, the generalization ability of our method is superior to prompt learning methods on novel sets(the last half of categories) of 11 classification datasets. Further comparisons of our method with prompt learning methods across the novel sets of 11 classification datasets are illustrated in \cref{tab:prompt learning methods on novel}. 
It's important to note that \cref{tab1:prompt learning methods} presents results on the base sets of these 11 classification datasets, conducted in the same 16-shot manner. When evaluating prompt learning methods on novel sets, they do not require additional images after fine-tuning the base categories of each dataset. Our method necessitates a few exemplar images to embed visual clues into the categories of novel sets.
Using just one image, our method surpasses current state-of-the-art (SoTA) methods in average performance across 11 datasets. With merely two images, our method achieves an unprecedented average performance exceeding 80.00\%. Increasing the number of images to 16 per category boosts average performance to 84.76\%, significantly outperforming current SoTA methods by 9.62\%. Considering the ease of collecting online data and our method's plug-and-play nature without extra training, it stands as an acceptable competitor to prompt learning methods in low-shot settings. Our method provides a generalizable and efficient approach to embedding multi-modal clues of novel classes into VLMs.

\begin{table*}[t]
\centering
\footnotesize
\setlength{\tabcolsep}{2.5pt}
\caption{\textbf{Open-vocabulary Classification Results on Novel Sets in Prompt Learning Setup.}}
\begin{tabular}{c|c|ccccccccccc|c}
\toprule
    Methods & Shot Number & ImageNet  & Caltech101 & OxfordPets & Cars & Flowers102 & Food101 & Aircraft & SUN397 & DTD & EuroSAT & UCF101 & Average \\
    \midrule
CLIP~\cite{CLIP}  & 0             & 68.14 & 94.00 & 97.26 & 74.89 & 77.80 & 91.22 & 36.29 & 75.35 & 59.90 & 64.05 & 77.50 & 74.22 \\
CoOp~\cite{coop}  & 0             & 67.88 & 89.81 & 95.29 & 60.40 & 59.67 & 82.26 & 22.30 & 65.89 & 41.18  & 54.74 & 56.05 & 63.22 \\
CoCoOp~\cite{cocoop} & 0         & 70.43 & 93.81 & 97.69 & 73.59 & 71.75 & 91.29 & 23.71 & 76.86 & 56.00 & 60.04 & 73.45 & 71.69 \\
MaPLe~\cite{khattak2023maple} & 0 & 70.54 & 94.36 & 97.76 & 74.00 & 72.46 & 92.05 & 35.61 & 78.70 & 59.18 & 73.23 & 78.66 & 75.14 \\ \hline
\multirow{4}{*}{Ours} & 1                 & 66.67 & 94.33 & 95.47 & 76.60 & 93.87 & 89.30 & 39.10 & 77.07 & 61.37 & 73.60 & 82.33 & 77.25 \\
                      & 2                 & 71.83 & 94.17 & 97.50 & 76.90 & 95.53 & 91.03 & 41.40 & 81.00 & 67.57 & 83.83 & 84.20 & 80.45 \\
                      & 4                 & 73.03 & 95.13 & 97.57 & 79.90 & 96.80 & 91.50 & 46.17 & 83.03 & 69.43 & 84.30 & 87.13 & 82.18 \\
                      & 16                & 74.87 & 96.30 & 97.67 & 86.23 & 97.13 & 91.70 & 52.03 & 84.60 & 74.73 & 89.57 & 87.57 & 84.76
\\ \bottomrule
\end{tabular}
\label{tab:prompt learning methods on novel}
\end{table*}

\section{Analysis of Preference Weight}

In this section, we delve into how the preference-based fusion module mitigates the adverse effects of low-quality text or images. This is achieved by adjusting the preference weights of different classifiers in response to the variable quality of multi-modal references. 

\begin{figure}[!h]
    \centering
    \includegraphics{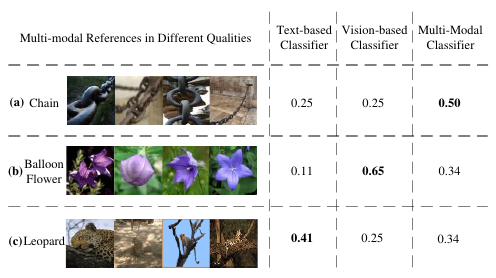}
    \caption{The variation in preference weight for different classifiers corresponding to multi-modal references of various qualities. (a) The category name ``Chain'' and the exemplar images are both of high quality and effectively complement each other. (b) The category name ``Balloon Flower'' may not describe the fine-grained flower in detail and is of low quality, whereas the exemplar images accurately represent the category. (c) The exemplar images with various backgrounds, poses, and appearances are of low quality, but the common word ``Leopard'' clearly defines the animal.}
    \label{fig:fusion_weight_analysis}
\end{figure}

In \cref{fig:fusion_weight_analysis}(a), both the category ``Chain'' and its corresponding exemplar images are of high quality or complementary to each other. Consequently, this category favors the multi-modal classifier, assigning it the highest preference weight of 0.50.
In \cref{fig:fusion_weight_analysis}(b), the category name ``Balloon Flower'' may fail to adequately describe the fine-grained characteristics of the flower in detail, reflecting its low quality, while the exemplar images depict the category more accurately. Thus, the vision-based classifier is assigned the highest preference weight, effectively mitigating the negative impact of the low-quality category name. Conversely, in \cref{fig:fusion_weight_analysis}(c) for the category ``Leopard'', the exemplar images are of low quality as a result of various backgrounds, poses, and appearances. In contrast, the common word ``Leopard'' can clearly illustrate the animal. Therefore, the text-based classifier receives the highest preference weight, compensating for the poor quality of the exemplar images.


\section{Sources of Exemplar Images}

\begin{figure*}[!h]
    \centering
    \includegraphics[width=0.95\linewidth]{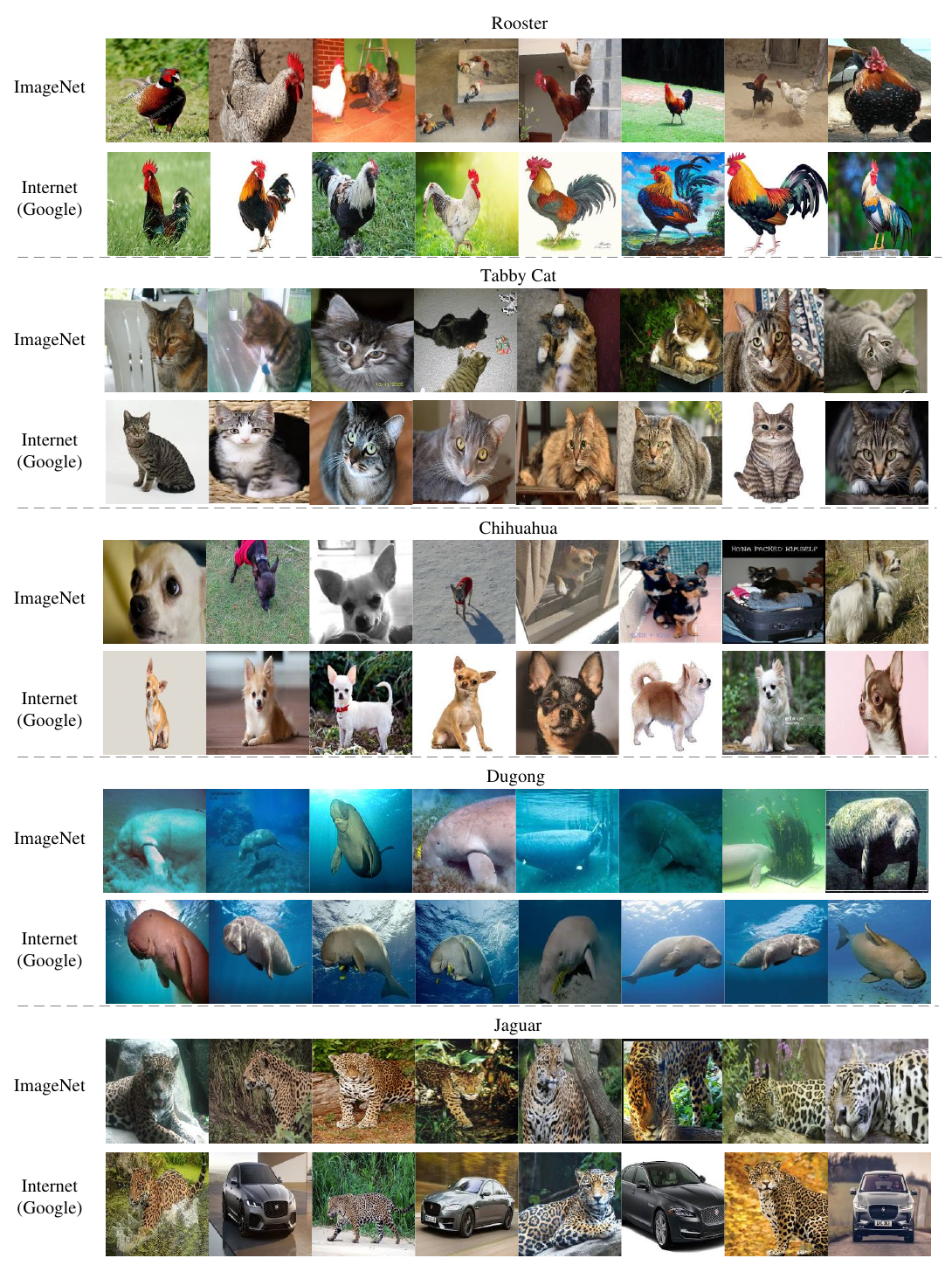}
    \caption{Exemplar images sampled from the training set of ImageNet and the Internet-crawled images.}
    \label{fig:img_sources}
\end{figure*}

In \cref{tab9:ablation_exemplar_source}, we showcase the performance of our method on the base classes of ImageNet using exemplar images sourced from the Internet and ImageNet's training set. When crawling images for a given category, we initiate the process by using the category name as a search query on Google. The first 16 images returned by Google are downloaded as the exemplar images for this category.
In \cref{fig:img_sources}, we present a set of examples crawled from the web and examples sampled from ImageNet's training set. 
It is evident that the images in ImageNet typically exhibit a higher diversity within the same class. The diversity includes differences in the background environment, the number of subjects, their poses, etc. Conversely, images sourced from the Internet often focus on a single subject, featuring simpler poses and backgrounds. Furthermore, images obtained from the Internet can be less reliable due to the noise and ambiguity inherent in text-based queries. For instance, a search for 'Jaguar' may yield images of either the animal or the car, as illustrated in the last row of \cref{fig:img_sources}. As demonstrated in \cref{tab9:ablation_exemplar_source}, the use of diverse images from ImageNet's training set as exemplars results in enhanced performance. This improvement is attributed to the closer domain correlation of the test set with the training set in ImageNet, as well as the potential noise and ambiguity of web-crawled images.

\section{Visualization of Detection Results}
\begin{figure*}[t]
    \centering
    \includegraphics[width=1.0\linewidth]{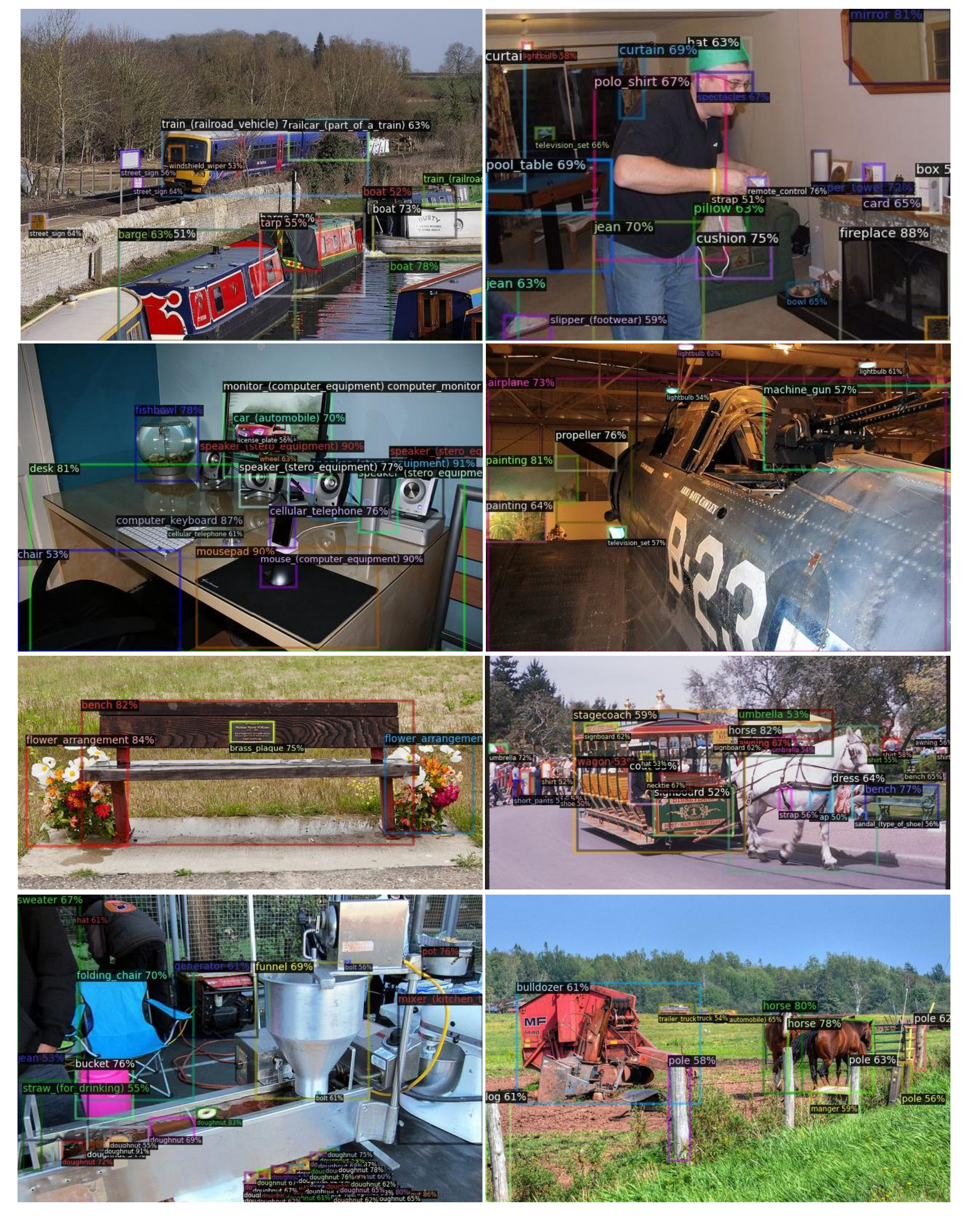}
    \caption{Detection results for all LVIS categories.}
    \label{fig:detection_all}
\end{figure*}

\begin{figure*}[t]
    \centering
    \includegraphics[width=1.0\linewidth]{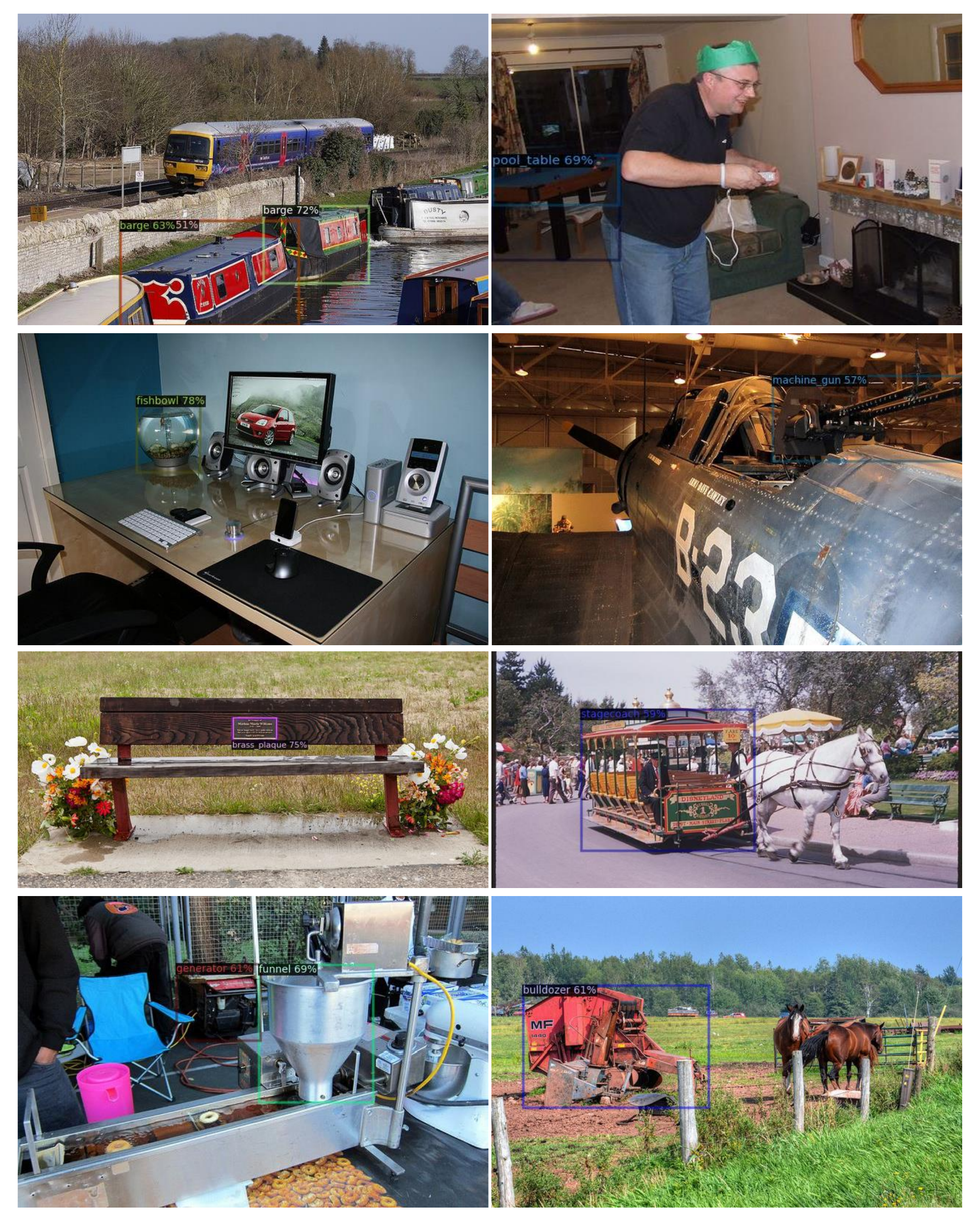}
    \caption{Detection results for novel LVIS categories.}
    \label{fig:detection_novel}
\end{figure*}

To demonstrate the effectiveness of our method in open-vocabulary detection, we separately showcase the detection results of our OVMR model using the Swin-B backbone, specifically for all categories in \cref{fig:detection_all} and for the novel categories in \cref{fig:detection_novel}, on the LVIS dataset.

%% file: main.bbl
\begin{thebibliography}{62}
\providecommand{\natexlab}[1]{#1}
\providecommand{\url}[1]{\texttt{#1}}
\expandafter\ifx\csname urlstyle\endcsname\relax
  \providecommand{\doi}[1]{doi: #1}\else
  \providecommand{\doi}{doi: \begingroup \urlstyle{rm}\Url}\fi

\bibitem[Bossard et~al.(2014)Bossard, Guillaumin, and
  Van~Gool]{bossard2014food}
Lukas Bossard, Matthieu Guillaumin, and Luc Van~Gool.
\newblock Food-101--mining discriminative components with random forests.
\newblock In \emph{ECCV}, 2014.

\bibitem[Chen et~al.(2019)Chen, Liu, Kira, Wang, and Huang]{chen2019closer}
Wei-Yu Chen, Yen-Cheng Liu, Zsolt Kira, Yu-Chiang~Frank Wang, and Jia-Bin
  Huang.
\newblock A closer look at few-shot classification.
\newblock \emph{arXiv preprint arXiv:1904.04232}, 2019.

\bibitem[Cimpoi et~al.(2014)Cimpoi, Maji, Kokkinos, Mohamed, and
  Vedaldi]{cimpoi2014describing}
Mircea Cimpoi, Subhransu Maji, Iasonas Kokkinos, Sammy Mohamed, and Andrea
  Vedaldi.
\newblock Describing textures in the wild.
\newblock In \emph{CVPR}, 2014.

\bibitem[Deng et~al.(2009)Deng, Dong, Socher, Li, Li, and
  Fei-Fei]{deng2009imagenet}
Jia Deng, Wei Dong, Richard Socher, Li-Jia Li, Kai Li, and Li Fei-Fei.
\newblock Imagenet: A large-scale hierarchical image database.
\newblock In \emph{CVPR}, 2009.

\bibitem[Du et~al.(2022{\natexlab{a}})Du, Wei, Zhang, Shi, Gao, and Li]{DetPro}
Yu Du, Fangyun Wei, Zihe Zhang, Miaojing Shi, Yue Gao, and Guoqi Li.
\newblock Learning to prompt for open-vocabulary object detection with
  vision-language model.
\newblock In \emph{CVPR}, 2022{\natexlab{a}}.

\bibitem[Du et~al.(2022{\natexlab{b}})Du, Wei, Zhang, Shi, Gao, and
  Li]{du2022learning}
Yu Du, Fangyun Wei, Zihe Zhang, Miaojing Shi, Yue Gao, and Guoqi Li.
\newblock Learning to prompt for open-vocabulary object detection with
  vision-language model.
\newblock In \emph{CVPR}, 2022{\natexlab{b}}.

\bibitem[Fei-Fei et~al.(2004)Fei-Fei, Fergus, and Perona]{fei2004learning}
Li Fei-Fei, Rob Fergus, and Pietro Perona.
\newblock Learning generative visual models from few training examples: An
  incremental bayesian approach tested on 101 object categories.
\newblock In \emph{CVPR-W}, 2004.

\bibitem[Feng et~al.(2022)Feng, Zhong, Jie, Chu, Ren, Wei, Xie, and
  Ma]{feng2022promptdet}
Chengjian Feng, Yujie Zhong, Zequn Jie, Xiangxiang Chu, Haibing Ren, Xiaolin
  Wei, Weidi Xie, and Lin Ma.
\newblock Promptdet: Towards open-vocabulary detection using uncurated images.
\newblock In \emph{European Conference on Computer Vision}, pages 701--717.
  Springer, 2022.

\bibitem[Finn et~al.(2017)Finn, Abbeel, and Levine]{finn2017model}
Chelsea Finn, Pieter Abbeel, and Sergey Levine.
\newblock Model-agnostic meta-learning for fast adaptation of deep networks.
\newblock In \emph{International conference on machine learning}, pages
  1126--1135. PMLR, 2017.

\bibitem[Gadre et~al.(2023)Gadre, Ilharco, Fang, Hayase, Smyrnis, Nguyen,
  Marten, Wortsman, Ghosh, Zhang, et~al.]{datacomp}
Samir~Yitzhak Gadre, Gabriel Ilharco, Alex Fang, Jonathan Hayase, Georgios
  Smyrnis, Thao Nguyen, Ryan Marten, Mitchell Wortsman, Dhruba Ghosh, Jieyu
  Zhang, et~al.
\newblock Datacomp: In search of the next generation of multimodal datasets.
\newblock \emph{arXiv preprint arXiv:2304.14108}, 2023.

\bibitem[Gao et~al.(2022)Gao, Xing, Niebles, Li, Xu, Liu, and Xiong]{pb-ovd}
Mingfei Gao, Chen Xing, Juan~Carlos Niebles, Junnan Li, Ran Xu, Wenhao Liu, and
  Caiming Xiong.
\newblock Open vocabulary object detection with pseudo bounding-box labels.
\newblock In \emph{ECCV}, 2022.

\bibitem[Gu et~al.(2022)Gu, Lin, Kuo, and Cui]{ViLD}
Xiuye Gu, Tsung-Yi Lin, Weicheng Kuo, and Yin Cui.
\newblock Open-vocabulary object detection via vision and language knowledge
  distillation.
\newblock In \emph{ICLR}, 2022.

\bibitem[Gupta et~al.(2019)Gupta, Dollar, and Girshick]{gupta2019lvis}
Agrim Gupta, Piotr Dollar, and Ross Girshick.
\newblock Lvis: A dataset for large vocabulary instance segmentation.
\newblock In \emph{Proceedings of the IEEE/CVF conference on computer vision
  and pattern recognition}, pages 5356--5364, 2019.

\bibitem[Han et~al.(2023)Han, Wei, Yu, Dou, He, Wang, Han, and
  Tian]{han2023boosting}
Xumeng Han, Longhui Wei, Xuehui Yu, Zhiyang Dou, Xin He, Kuiran Wang, Zhenjun
  Han, and Qi Tian.
\newblock Boosting segment anything model towards open-vocabulary learning.
\newblock \emph{arXiv preprint arXiv:2312.03628}, 2023.

\bibitem[Hao et~al.(2023)Hao, He, Liu, Wu, Tao, and Cheng]{Hao_2023_ICCV}
Fusheng Hao, Fengxiang He, Liu Liu, Fuxiang Wu, Dacheng Tao, and Jun Cheng.
\newblock Class-aware patch embedding adaptation for few-shot image
  classification.
\newblock In \emph{Proceedings of the IEEE/CVF International Conference on
  Computer Vision (ICCV)}, pages 18905--18915, 2023.

\bibitem[Helber et~al.(2019)Helber, Bischke, Dengel, and
  Borth]{helber2019eurosat}
Patrick Helber, Benjamin Bischke, Andreas Dengel, and Damian Borth.
\newblock Eurosat: A novel dataset and deep learning benchmark for land use and
  land cover classification.
\newblock \emph{IEEE Journal of Selected Topics in Applied Earth Observations
  and Remote Sensing}, 2019.

\bibitem[Kaul et~al.(2023)Kaul, Xie, and Zisserman]{kaul2023multi}
Prannay Kaul, Weidi Xie, and Andrew Zisserman.
\newblock Multi-modal classifiers for open-vocabulary object detection.
\newblock \emph{arXiv preprint arXiv:2306.05493}, 2023.

\bibitem[Khattak et~al.(2023)Khattak, Rasheed, Maaz, Khan, and
  Khan]{khattak2023maple}
Muhammad~Uzair Khattak, Hanoona Rasheed, Muhammad Maaz, Salman Khan, and
  Fahad~Shahbaz Khan.
\newblock Maple: Multi-modal prompt learning.
\newblock In \emph{Proceedings of the IEEE/CVF Conference on Computer Vision
  and Pattern Recognition}, pages 19113--19122, 2023.

\bibitem[Kim et~al.(2023{\natexlab{a}})Kim, Angelova, and
  Kuo]{kim2023contrastive}
Dahun Kim, Anelia Angelova, and Weicheng Kuo.
\newblock Contrastive feature masking open-vocabulary vision transformer.
\newblock In \emph{Proceedings of the IEEE/CVF International Conference on
  Computer Vision}, pages 15602--15612, 2023{\natexlab{a}}.

\bibitem[Kim et~al.(2023{\natexlab{b}})Kim, Angelova, and
  Kuo]{kim2023detection}
Dahun Kim, Anelia Angelova, and Weicheng Kuo.
\newblock Detection-oriented image-text pretraining for open-vocabulary
  detection.
\newblock \emph{arXiv preprint arXiv:2310.00161}, 2023{\natexlab{b}}.

\bibitem[Kim et~al.(2023{\natexlab{c}})Kim, Angelova, and Kuo]{rovit}
Dahun Kim, Anelia Angelova, and Weicheng Kuo.
\newblock Region-aware pretraining for open-vocabulary object detection with
  vision transformers.
\newblock In \emph{Conference on Computer Vision and Pattern Recognition
  (CVPR)}, 2023{\natexlab{c}}.

\bibitem[Krause et~al.(2013)Krause, Stark, Deng, and Fei-Fei]{krause20133d}
Jonathan Krause, Michael Stark, Jia Deng, and Li Fei-Fei.
\newblock 3d object representations for fine-grained categorization.
\newblock In \emph{ICCV-W}, 2013.

\bibitem[Kuo et~al.(2023)Kuo, Cui, Gu, Piergiovanni, and Angelova]{kuo2022fvlm}
Weicheng Kuo, Yin Cui, Xiuye Gu, AJ Piergiovanni, and Anelia Angelova.
\newblock F-vlm: Open-vocabulary object detection upon frozen vision and
  language models.
\newblock \emph{ICLR}, 2023.

\bibitem[Li et~al.(2022{\natexlab{a}})Li, He, Wei, Qian, Zhu, Xie, Zhuang,
  Tian, and Tang]{li2022fine}
Juncheng Li, Xin He, Longhui Wei, Long Qian, Linchao Zhu, Lingxi Xie, Yueting
  Zhuang, Qi Tian, and Siliang Tang.
\newblock Fine-grained semantically aligned vision-language pre-training.
\newblock \emph{Advances in neural information processing systems},
  35:\penalty0 7290--7303, 2022{\natexlab{a}}.

\bibitem[Li et~al.(2022{\natexlab{b}})Li, Zhang, Zhang, Yang, Li, Zhong, Wang,
  Yuan, Zhang, Hwang, Chang, and Gao]{GLIP}
Liunian~Harold Li, Pengchuan Zhang, Haotian Zhang, Jianwei Yang, Chunyuan Li,
  Yiwu Zhong, Lijuan Wang, Lu Yuan, Lei Zhang, Jenq-Neng Hwang, Kai-Wei Chang,
  and Jianfeng Gao.
\newblock Grounded language-image pre-training.
\newblock In \emph{CVPR}, 2022{\natexlab{b}}.

\bibitem[Li et~al.(2021)Li, Liu, and Bilen]{li2021universal}
Wei-Hong Li, Xialei Liu, and Hakan Bilen.
\newblock Universal representation learning from multiple domains for few-shot
  classification.
\newblock In \emph{Proceedings of the IEEE/CVF International Conference on
  Computer Vision}, pages 9526--9535, 2021.

\bibitem[Li et~al.(2023)Li, Fan, Hu, Feichtenhofer, and He]{FLIP}
Yanghao Li, Haoqi Fan, Ronghang Hu, Christoph Feichtenhofer, and Kaiming He.
\newblock Scaling language-image pre-training via masking.
\newblock In \emph{Proceedings of the IEEE/CVF Conference on Computer Vision
  and Pattern Recognition}, pages 23390--23400, 2023.

\bibitem[Liu et~al.(2023)Liu, Li, Wu, and Lee]{liu2023llava}
Haotian Liu, Chunyuan Li, Qingyang Wu, and Yong~Jae Lee.
\newblock Visual instruction tuning.
\newblock In \emph{NeurIPS}, 2023.

\bibitem[Luo et~al.(2023)Luo, Wu, Zhang, Gao, Xu, and Song]{luo2023few-shot}
Xu Luo, Hao Wu, Ji Zhang, Lianli Gao, Jing Xu, and Jingkuan Song.
\newblock A closer look at few-shot classification again.
\newblock In \emph{Proceedings of the 40th International Conference on Machine
  Learning}. JMLR.org, 2023.

\bibitem[Ma et~al.(2023)Ma, Jiang, Wen, Yuan, and Qi]{ma2023codet}
Chuofan Ma, Yi Jiang, Xin Wen, Zehuan Yuan, and Xiaojuan Qi.
\newblock Codet: Co-occurrence guided region-word alignment for open-vocabulary
  object detection.
\newblock \emph{arXiv preprint arXiv:2310.16667}, 2023.

\bibitem[Maji et~al.(2013)Maji, Rahtu, Kannala, Blaschko, and
  Vedaldi]{maji2013fine}
Subhransu Maji, Esa Rahtu, Juho Kannala, Matthew Blaschko, and Andrea Vedaldi.
\newblock Fine-grained visual classification of aircraft.
\newblock \emph{arXiv preprint arXiv:1306.5151}, 2013.

\bibitem[Minderer et~al.(2022)Minderer, Gritsenko, Stone, Neumann, Weissenborn,
  Dosovitskiy, Mahendran, Arnab, Dehghani, Shen, Wang, Zhai, Kipf, and
  Houlsby]{minderer2022simple}
Matthias Minderer, Alexey Gritsenko, Austin Stone, Maxim Neumann, Dirk
  Weissenborn, Alexey Dosovitskiy, Aravindh Mahendran, Anurag Arnab, Mostafa
  Dehghani, Zhuoran Shen, Xiao Wang, Xiaohua Zhai, Thomas Kipf, and Neil
  Houlsby.
\newblock Simple open-vocabulary object detection with vision transformers.
\newblock In \emph{ECCV}, 2022.

\bibitem[Ni et~al.(2023)Ni, Wei, Tang, Zhuang, and Tian]{ni2023continual}
Zixuan Ni, Longhui Wei, Siliang Tang, Yueting Zhuang, and Qi Tian.
\newblock Continual vision-language representation learning with off-diagonal
  information.
\newblock In \emph{International Conference on Machine Learning}, pages
  26129--26149. PMLR, 2023.

\bibitem[Nilsback and Zisserman(2008)]{nilsback2008automated}
Maria-Elena Nilsback and Andrew Zisserman.
\newblock Automated flower classification over a large number of classes.
\newblock In \emph{ICVGIP}, 2008.

\bibitem[Parkhi et~al.(2012)Parkhi, Vedaldi, Zisserman, and
  Jawahar]{parkhi2012cats}
Omkar~M Parkhi, Andrea Vedaldi, Andrew Zisserman, and CV Jawahar.
\newblock Cats and dogs.
\newblock In \emph{CVPR}, 2012.

\bibitem[Radford et~al.(2021)Radford, Kim, Hallacy, Ramesh, Goh, Agarwal,
  Sastry, Askell, Mishkin, Clark, et~al.]{CLIP}
Alec Radford, Jong~Wook Kim, Chris Hallacy, Aditya Ramesh, Gabriel Goh,
  Sandhini Agarwal, Girish Sastry, Amanda Askell, Pamela Mishkin, Jack Clark,
  et~al.
\newblock Learning transferable visual models from natural language
  supervision.
\newblock In \emph{International conference on machine learning}, pages
  8748--8763. PMLR, 2021.

\bibitem[Schuhmann et~al.(2022)Schuhmann, Beaumont, Vencu, Gordon, Wightman,
  Cherti, Coombes, Katta, Mullis, Wortsman, Schramowski, Kundurthy, Crowson,
  Schmidt, Kaczmarczyk, and Jitsev]{laion5b}
Christoph Schuhmann, Romain Beaumont, Richard Vencu, Cade~W Gordon, Ross
  Wightman, Mehdi Cherti, Theo Coombes, Aarush Katta, Clayton Mullis, Mitchell
  Wortsman, Patrick Schramowski, Srivatsa~R Kundurthy, Katherine Crowson,
  Ludwig Schmidt, Robert Kaczmarczyk, and Jenia Jitsev.
\newblock {LAION}-5b: An open large-scale dataset for training next generation
  image-text models.
\newblock In \emph{Thirty-sixth Conference on Neural Information Processing
  Systems Datasets and Benchmarks Track}, 2022.

\bibitem[Snell et~al.(2017)Snell, Swersky, and Zemel]{snell2017prototypical}
Jake Snell, Kevin Swersky, and Richard Zemel.
\newblock Prototypical networks for few-shot learning.
\newblock \emph{Advances in neural information processing systems}, 30, 2017.

\bibitem[Sokolova and Lapalme(2009)]{sokolova2009systematic}
Marina Sokolova and Guy Lapalme.
\newblock A systematic analysis of performance measures for classification
  tasks.
\newblock \emph{Information processing \& management}, 45\penalty0
  (4):\penalty0 427--437, 2009.

\bibitem[Soomro et~al.(2012)Soomro, Zamir, and Shah]{soomro2012ucf101}
Khurram Soomro, Amir~Roshan Zamir, and Mubarak Shah.
\newblock Ucf101: A dataset of 101 human actions classes from videos in the
  wild.
\newblock \emph{arXiv preprint arXiv:1212.0402}, 2012.

\bibitem[Sun et~al.(2023)Sun, Fang, Wu, Wang, and Cao]{sun2023eva}
Quan Sun, Yuxin Fang, Ledell Wu, Xinlong Wang, and Yue Cao.
\newblock Eva-clip: Improved training techniques for clip at scale.
\newblock \emph{arXiv preprint arXiv:2303.15389}, 2023.

\bibitem[Tian et~al.(2020)Tian, Wang, Krishnan, Tenenbaum, and
  Isola]{tian2020rethinking}
Yonglong Tian, Yue Wang, Dilip Krishnan, Joshua~B Tenenbaum, and Phillip Isola.
\newblock Rethinking few-shot image classification: a good embedding is all you
  need?
\newblock In \emph{Computer Vision--ECCV 2020: 16th European Conference,
  Glasgow, UK, August 23--28, 2020, Proceedings, Part XIV 16}, pages 266--282.
  Springer, 2020.

\bibitem[Vinyals et~al.(2016)Vinyals, Blundell, Lillicrap, Wierstra,
  et~al.]{vinyals2016matching}
Oriol Vinyals, Charles Blundell, Timothy Lillicrap, Daan Wierstra, et~al.
\newblock Matching networks for one shot learning.
\newblock \emph{Advances in neural information processing systems}, 29, 2016.

\bibitem[Wang and Zhang(2023)]{wang2023contextual}
Dongkai Wang and Shiliang Zhang.
\newblock Contextual instance decoupling for instance-level human analysis.
\newblock \emph{IEEE Transactions on Pattern Analysis and Machine
  Intelligence}, 2023.

\bibitem[Wang(2023)]{wang2023learning}
Tao Wang.
\newblock Learning to detect and segment for open vocabulary object detection.
\newblock In \emph{Proceedings of the IEEE/CVF Conference on Computer Vision
  and Pattern Recognition}, pages 7051--7060, 2023.

\bibitem[Wei et~al.(2022)Wei, Xie, Zhou, Li, and Tian]{wei2022mvp}
Longhui Wei, Lingxi Xie, Wengang Zhou, Houqiang Li, and Qi Tian.
\newblock Mvp: Multimodality-guided visual pre-training.
\newblock In \emph{European Conference on Computer Vision}, pages 337--353.
  Springer, 2022.

\bibitem[Wu et~al.(2023{\natexlab{a}})Wu, Zhang, Jin, Liu, and
  Loy]{wu2023aligning}
Size Wu, Wenwei Zhang, Sheng Jin, Wentao Liu, and Chen~Change Loy.
\newblock Aligning bag of regions for open-vocabulary object detection.
\newblock In \emph{Proceedings of the IEEE/CVF Conference on Computer Vision
  and Pattern Recognition}, pages 15254--15264, 2023{\natexlab{a}}.

\bibitem[Wu et~al.(2023{\natexlab{b}})Wu, Zhu, Zhao, and Li]{wu2023cora}
Xiaoshi Wu, Feng Zhu, Rui Zhao, and Hongsheng Li.
\newblock Cora: Adapting clip for open-vocabulary detection with region
  prompting and anchor pre-matching.
\newblock In \emph{Proceedings of the IEEE/CVF Conference on Computer Vision
  and Pattern Recognition}, pages 7031--7040, 2023{\natexlab{b}}.

\bibitem[Xiao et~al.(2010)Xiao, Hays, Ehinger, Oliva, and
  Torralba]{xiao2010sun}
Jianxiong Xiao, James Hays, Krista~A Ehinger, Aude Oliva, and Antonio Torralba.
\newblock Sun database: Large-scale scene recognition from abbey to zoo.
\newblock In \emph{CVPR}, 2010.

\bibitem[Xu et~al.(2023)Xu, Zhang, Fu, Chen, Yang, Li, and Xu]{xu2023multi}
Yifan Xu, Mengdan Zhang, Chaoyou Fu, Peixian Chen, Xiaoshan Yang, Ke Li, and
  Changsheng Xu.
\newblock Multi-modal queried object detection in the wild.
\newblock \emph{arXiv preprint arXiv:2305.18980}, 2023.

\bibitem[Xuan et~al.(2023)Xuan, Guo, Yang, and Zhang]{xuan2023pink}
Shiyu Xuan, Qingpei Guo, Ming Yang, and Shiliang Zhang.
\newblock Pink: Unveiling the power of referential comprehension for
  multi-modal llms.
\newblock \emph{arXiv preprint arXiv:2310.00582}, 2023.

\bibitem[Yao et~al.(2022)Yao, Han, Wen, Liang, Xu, Zhang, Li, Xu, , and
  Xu]{detclip}
Lewei Yao, Jianhua Han, Youpeng Wen, Xiaodan Liang, Dan Xu, Wei Zhang, Zhenguo
  Li, Chunjing Xu, , and Hang Xu.
\newblock Detclip: Dictionary-enriched visual-concept paralleled pre-training
  for open-world detection.
\newblock In \emph{arXiv:2209.09407}, 2022.

\bibitem[Yao et~al.(2023)Yao, Han, Liang, Xu, Zhang, Li, and
  Xu]{yao2023detclipv2}
Lewei Yao, Jianhua Han, Xiaodan Liang, Dan Xu, Wei Zhang, Zhenguo Li, and Hang
  Xu.
\newblock Detclipv2: Scalable open-vocabulary object detection pre-training via
  word-region alignment.
\newblock In \emph{Proceedings of the IEEE/CVF Conference on Computer Vision
  and Pattern Recognition}, pages 23497--23506, 2023.

\bibitem[Zang et~al.(2022)Zang, Li, Zhou, Huang, and Loy]{zang2022open}
Yuhang Zang, Wei Li, Kaiyang Zhou, Chen Huang, and Chen~Change Loy.
\newblock Open-vocabulary {D}{E}{T}{R} with conditional matching.
\newblock In \emph{ECCV}, pages 106--122. Springer, 2022.

\bibitem[Zareian et~al.(2021)Zareian, Rosa, Hu, and Chang]{Zareian_2021_CVPR}
Alireza Zareian, Kevin~Dela Rosa, Derek~Hao Hu, and Shih-Fu Chang.
\newblock Open-vocabulary object detection using captions.
\newblock In \emph{CVPR}, 2021.

\bibitem[Zhang et~al.(2022)Zhang, Zhang, Hu, Chen, Li, Dai, Wang, Yuan, Hwang,
  and Gao]{zhang2022glipv2}
Haotian Zhang, Pengchuan Zhang, Xiaowei Hu, Yen-Chun Chen, Liunian Li, Xiyang
  Dai, Lijuan Wang, Lu Yuan, Jenq-Neng Hwang, and Jianfeng Gao.
\newblock Glipv2: Unifying localization and vision-language understanding.
\newblock \emph{Advances in Neural Information Processing Systems},
  35:\penalty0 36067--36080, 2022.

\bibitem[Zhong et~al.(2022)Zhong, Yang, Zhang, Li, Codella, Li, Zhou, Dai,
  Yuan, Li, and Gao]{zhong2021regionclip}
Yiwu Zhong, Jianwei Yang, Pengchuan Zhang, Chunyuan Li, Noel Codella,
  Liunian~Harold Li, Luowei Zhou, Xiyang Dai, Lu Yuan, Yin Li, and Jianfeng
  Gao.
\newblock Regionclip: Region-based language-image pretraining.
\newblock In \emph{CVPR}, 2022.

\bibitem[Zhou et~al.(2022{\natexlab{a}})Zhou, Yang, Loy, and Liu]{cocoop}
Kaiyang Zhou, Jingkang Yang, Chen~Change Loy, and Ziwei Liu.
\newblock Conditional prompt learning for vision-language models.
\newblock In \emph{Proceedings of the IEEE/CVF Conference on Computer Vision
  and Pattern Recognition}, pages 16816--16825, 2022{\natexlab{a}}.

\bibitem[Zhou et~al.(2022{\natexlab{b}})Zhou, Yang, Loy, and Liu]{coop}
Kaiyang Zhou, Jingkang Yang, Chen~Change Loy, and Ziwei Liu.
\newblock Learning to prompt for vision-language models.
\newblock \emph{International Journal of Computer Vision}, 130\penalty0
  (9):\penalty0 2337--2348, 2022{\natexlab{b}}.

\bibitem[Zhou et~al.(2019)Zhou, Wang, and Kr{\"a}henb{\"u}hl]{zhou2019objects}
Xingyi Zhou, Dequan Wang, and Philipp Kr{\"a}henb{\"u}hl.
\newblock Objects as points.
\newblock \emph{arXiv preprint arXiv:1904.07850}, 2019.

\bibitem[Zhou et~al.(2021)Zhou, Koltun, and
  Kr{\"a}henb{\"u}hl]{zhou2021probabilistic}
Xingyi Zhou, Vladlen Koltun, and Philipp Kr{\"a}henb{\"u}hl.
\newblock Probabilistic two-stage detection.
\newblock \emph{arXiv preprint arXiv:2103.07461}, 2021.

\bibitem[Zhou et~al.(2022{\natexlab{c}})Zhou, Girdhar, Joulin,
  Kr{\"a}henb{\"u}hl, and Misra]{zhou2022detecting}
Xingyi Zhou, Rohit Girdhar, Armand Joulin, Philipp Kr{\"a}henb{\"u}hl, and
  Ishan Misra.
\newblock Detecting twenty-thousand classes using image-level supervision.
\newblock In \emph{ECCV}, 2022{\natexlab{c}}.

\end{thebibliography}
